\documentclass[11pt,oneside,a4paper]{article}
\usepackage[top=1 in, bottom=1 in, left=0.85 in, right=0.85 in]{geometry}
\usepackage[hidelinks]{hyperref}
\usepackage{multirow}
\usepackage[table,xcdraw]{xcolor}
\usepackage{amsmath,amssymb,graphicx,url, algorithm2e}
\usepackage{thmtools,thm-restate,wrapfig,enumitem,mathabx}
\usepackage{float}
\usepackage{graphicx}
\usepackage{subcaption}
\usepackage{microtype}
\newtheorem{assumption}{Assumption}
\newtheorem{theorem}{Theorem}
\newtheorem{condition}{Condition}
\newtheorem{remark}{Remark}
\newtheorem{lemma}{Lemma}

\usepackage{makecell}

\DeclareMathOperator*{\argmax}{arg\,max}

\title{Rectified Pessimistic-Optimistic Learning for Stochastic Continuum-armed Bandit with Constraints}
\author{Hengquan Guo, Qi Zhu, and Xin Liu \\ShanghaiTech University \\ \{guohq, zhuqi2022,liuxin7\}@shanghaitech.edu.cn}
\date{}

\begin{document}

\maketitle

\begin{abstract}
This paper studies the problem of stochastic continuum-armed bandit with constraints (SCBwC), where we optimize a black-box reward function $f(x)$ subject to a black-box constraint function $g(x)\leq 0$ over a continuous space $\mathcal X$. We model reward and constraint functions via Gaussian processes (GPs) and propose a Rectified Pessimistic-Optimistic Learning framework (RPOL), a penalty-based method incorporating optimistic and pessimistic GP bandit learning for reward and constraint functions, respectively. 
We consider the metric of cumulative constraint violation $\sum_{t=1}^T(g(x_t))^{+},$ which is strictly stronger than the traditional long-term constraint violation $\sum_{t=1}^Tg(x_t).$ 
The rectified design for the penalty update and the pessimistic learning for the constraint function in RPOL guarantee the cumulative constraint violation is minimal. RPOL can achieve sublinear regret and cumulative constraint violation for SCBwC and its variants (e.g., under delayed feedback and non-stationary environment). These theoretical results match their unconstrained counterparts. Our experiments justify RPOL outperforms several existing baseline algorithms.   
\end{abstract}

\section{Introduction}
Stochastic continuum-armed bandit optimization is a powerful framework to model many real-world applications, (e.g., networking resource allocation \cite{FuMod_21}, online recommendation \cite{KraOng_11}, clinic trials \cite{DurAchIac_18}, neural network architecture search \cite{WhiNeiSav_21}. In stochastic continuum-armed bandits, the learner aims to optimize a black-box reward/utility function over a continuous feasible set $\mathcal X$ by sequentially interacting with the environment. The interaction with the practical environment is often subject to a variety of operational constraints, which are also black-box and complicated. For example, in networking resource allocation, we maximize the users' quality of experience under complex resource constraints; in clinic trials, we optimize the quality of treatment while guaranteeing the side effect of patients minimal; in the neural architecture search, we search a neural network with a small generalization error while keeping the training time within the time limit. In these applications, the learner requires to optimize a black-box reward/utility function $f(x)$ while keeping the black-box constraint $(g(x) \leq 0)$ satisfied. 
The black-box problem is unsolvable in general without any regularity assumption on $f(x)$ and $g(x).$ We assume the reward and constraint functions lie in Reproducing Kernel Hilbert Space (RKHS) with a bounded norm such that $f(x)$ and $g(x)$ can be modeled via Gaussian processes (GPs).   

The previous works in stochastic continuum-armed bandit with constraints (SCBwC) are classified into two categories according to the type of constraints: hard and soft constraints, respectively. For the type of hard constraints, there is a sequence of studies on safe Bayesian optimization \cite{AmaAliThr_20,BerKraSch_21,SuiGotBur_15,SuiZhuBur_18}, where the algorithms satisfy the constraint instantaneously at each round, i.e., hard constraint. However, these results rely on the key assumption that an initial safe/feasible decision set is known apriori; otherwise, it would be impossible to guarantee the hard constraints. Moreover, the algorithms in \cite{AmaAliThr_20,BerKraSch_21,SuiGotBur_15,SuiZhuBur_18} suffer from high-computation complexity because they require to construct a safe decision set and search for a safe and optimal solution for each round. Without any prior information on the constraint function or safe set, the constraint violation is {\it unavoidable}.  A recent line of work focuses on the soft constraints \cite{AriColBro_19,ShiEry_22,ZhouJi_22}, which allow the constraints to be violated as long as they are satisfied in the long term. In other words, the soft constraint violation $\sum_{t=1}^T g(x_t)$ should be as small as possible. The soft constraint violation is a reasonable metric for the long-term budget or fairness constraints. However, it is improper for safety-critical applications because we may have a sequence of decisions with zero soft constraint violation and violates the constraints at every round. For example, consider a sequence of decisions $\{g_{t}(x_t)\}$ such that $g_{t}(x_t) = -1$ if $t$ is odd and $g_{t}(x_t) = +1$ if $t$ is even. For such a sequence with $T=1000,$ we have $\sum_{t=1}^\tau g_t(x_t)\leq 0$ for any $1\leq \tau \leq T,$ but the constraint violates at half of $T$ rounds.

In this paper, we focus on stochastic continuum-armed bandit with constraints (SCBwC) via the Gaussian processes model and study the cumulative constraint violation $\sum_{t=1}^T (g(x_t))^{+}.$ The cumulative violation is a strictly stronger metric than the soft violation because it cannot be compensated among different rounds.   
Our goal is to optimize a black-box reward function while keeping the cumulative violation minimal. In this paper, we propose a Rectified Pessimistic-Optimistic Learning (RPOL), an efficient penalty-based framework integrating optimistic and pessimistic estimators of reward and constraint functions into a single surrogate function. The framework acquires the information of block-box reward and constraint functions efficiently and safely, and it is flexible to achieve strong performance in SCBwC and its variants (bandits with delayed feedback or bandits under non-stationary environment). It is worth to be emphasizing that a concurrent work \cite{XuJiaJon_22} also considers the cumulative violation. However, it requires solving a complex constrained optimization problem for each round that might suffer from high computational complexity, and it is not clear if their method can be applied to bandits with delayed feedback or non-stationary bandits as in our paper. Moreover, our experiments show RPOL outperforms their method w.r.t. both reward and constraint violation.

\subsection{Main Contribution}
{\bf Algorithm Design}~This paper proposes a rectified pessimistic-optimistic learning framework (RPOL) for SCBwC, where the rectified design is to avoid aggressive exploration and encourages conservative/pessimistic decisions such that it can minimize the cumulative constraint violation.  
The proposed framework is flexible to incorporate the classical exploration strategies in Gaussian process bandit learning (e.g., GP-UCB in \cite{SriKraKak_09} or improved GP-UCB in \cite{ChoSayGop_17}) and provides the strong performance guarantee in regret and cumulative violation. Moreover, our framework is also readily applied to the variants of SCBwC, (e.g., bandits with delayed feedback  in Section \ref{sec:delay} and bandits under non-stationary environment in Section \ref{sec:non-stationary}).

\vspace{5pt}
\noindent{\bf Theoretical Results}~We develop a unified analysis method for RPOL framework in Theorem \ref{thm: main}, where the regret and cumulative violation depend on the errors of optimistic or pessimistic learning. The method is quite general to be used in analyzing SCBwC and its variants, and we establish the following theoretical results ($\gamma_T$ is the information gain w.r.t. the kernel used to approximate reward and constraint functions via GPs). 
\begin{itemize}[leftmargin=2em]
    \item For SCBwC, we instantiate RPOL with GP-UCB (RPOL-UCB) and prove it achieves $O\left(\gamma_T\sqrt{T}\right)$ regret and cumulative constraint violation. RPOL-UCB strictly improves \cite{ZhouJi_22} as shown in Table \ref{tab:main contribution} and achieves similar performance with an efficient penalty-based method compared to the concurrent work \cite{XuJiaJon_22}, a constrained optimization-based method. 
    
    \item For SCBwC with delayed feedback, we integrate RPOL with censored GP-UCB (RPOL-CensoredUCB) and show it achieves $O\left(\frac{\gamma_T}{\rho_m}(\sqrt{T} + m)\right)$ regret and cumulative violation, where $m$ and $\rho_m$ are the parameters related to the delay, as shown in Table \ref{tab:main contribution-delay}. To the best of our knowledge, this is the first result in SCBwC with delayed feedback.

    \item For SCBwC under non-stationary environment, we instantiate RPOL with sliding window GP-UCB  (RPOL-SWUCB) and show it achieves $O\left(\gamma_T^{7/8} P_T^{1/4}T^{3/4}\right)$ regret and cumulative violation as shown in Table \ref{tab:main contribution-non-stationary}, where $P_T$ is the total variation of reward and constraint function. To the best of our knowledge, this is also the first result in SCBwC under non-stationary environment.
\end{itemize}

\begin{table*}
\centering
\begin{tabular}{|c|c|c|c|c|}
\hline
Reference                            & Regret                                          & Soft Violation                                                  & Hard Violation  & Design Method   \\ \hline
\cite{ZhouJi_22} &$O(\gamma_T\sqrt{T})$ &$O(\gamma_T \sqrt{T}/ \chi)$ &N/A & Primal-dual\\ \cline{1-1} \cline{2-3} \cline{3-4} \cline{4-5} 
 \cline{1-1} \cline{2-3} \cline{3-4} \cline{4-5}
\cite{XuJiaJon_22}&$O(\gamma_T\sqrt{T})$&$O(\gamma_T\sqrt{T})$&$O(\gamma_T\sqrt{T})$& Constrained optimization
\\ 
 \cline{1-1} \cline{2-3} \cline{3-4} \cline{4-5}
 \cellcolor[HTML]{C0C0C0} RPOL-UCB &$\cellcolor[HTML]{C0C0C0}O(\gamma_T\sqrt{T})$                                                       & \cellcolor[HTML]{C0C0C0}$O(\gamma_T\sqrt{T})$  &\cellcolor[HTML]{C0C0C0}$O(\gamma_T\sqrt{T})$                   & \cellcolor[HTML]{C0C0C0}Penalty\\ \cline{1-1} \cline{2-3} \cline{3-4} \cline{4-5}
\end{tabular}
\caption{Our results and related work in SCBwC, where $\chi$ is a constant related to Slater's condition of the offline problem in \eqref{def: obj}-\eqref{def: cons} and requires to be known in \cite{ZhouJi_22}. \cite{ZhouJi_22} and this paper can be regarded as unconstrained optimization methods, and \cite{XuJiaJon_22} is a constrained optimization-based method.
}
\label{tab:main contribution}
\end{table*}

\begin{table*}
\centering
\begin{tabular}{|c|c|c|}
\hline
Reference                            & Regret                                          & Hard Violation                                                  \\ \hline
\cite{VerDaiLow_22} &$O\left(\frac{\gamma_T}{\rho_m}(\sqrt{T} + m)\right)$ &N/A  \\ \cline{1-1} \cline{2-3}  
 \cellcolor[HTML]{C0C0C0} RPOL-CensoredUCB &$\cellcolor[HTML]{C0C0C0}O\left(\frac{\gamma_T}{\rho_m}(\sqrt{T} + m)\right) $                                                  &$\cellcolor[HTML]{C0C0C0}O\left(\frac{\gamma_T}{\rho_m}(\sqrt{T} + m)\right) $                      \\ \cline{1-1} \cline{2-3}
\end{tabular}
\caption{Our results and related work in SCBwC under delayed feedback.}
\label{tab:main contribution-delay}
\end{table*}

\begin{table*}
\centering
\begin{tabular}{|c|c|c|c|}
\hline
Reference                            & Regret                &Soft Violation                          & Hard Violation                                                  \\ \hline
\cite{DenZhoGho_22} &$O\left(\chi\gamma_T^{7/8} P_T^{1/4}T^{3/4} \right)$ &$O\left((1+ \frac{1}{\chi})\gamma_T^{7/8} P_T^{1/4}T^{3/4} \right)$ &N/A  \\ \cline{1-1} \cline{2-3}  \cline{3-4}
 \cellcolor[HTML]{C0C0C0} RPOL-SWUCB &$\cellcolor[HTML]{C0C0C0}O\left(\gamma_T^{7/8} P_T^{1/4}T^{3/4}\right) $     &$\cellcolor[HTML]{C0C0C0}O\left(\gamma_T^{7/8} P_T^{1/4}T^{3/4}\right) $                                            &$\cellcolor[HTML]{C0C0C0}O\left(\gamma_T^{7/8} P_T^{1/4}T^{3/4}\right) $                      \\ \cline{1-1} \cline{2-3}\cline{3-4}
\end{tabular}
\caption{Our results and related work in SCBwC under non-stationary environment.}
\label{tab:main contribution-non-stationary}
\end{table*}

\subsection{Related Work}

{\bf Stochastic Continuum-armed Bandit with Constraints} 
The stochastic continuum-armed bandit with constraints is widely used to model safety-critical applications (e.g., \cite{AmaAliThr_20, BerKraSch_21, SuiGotBur_15, SuiZhuBur_18}), where safety constraints are imposed and required to be satisfied instantaneously. These works assume an initial safe decision set and establish 
and the algorithm would suffer from high computation complexity and suboptimal performance due to overly conservative decisions. The work \cite{ShiEry_22} and \cite{ZhouJi_22} studied constrained kernelized bandits with long-term constraints with the metric of soft constraint violation of $\sum_{t=1}^T g(x_t).$ The work \cite{ShiEry_22} proposed a penalty-based algorithm, which  achieves $O(\gamma_{T}T^{3/4})$ regret and  $O(T^{3/4})$ violation; and the work \cite{ZhouJi_22} proposed a primal-dual algorithm and achieved $O(\gamma_T\sqrt{T})$ regret and $O(\gamma_T(1+\frac{1}{\chi})\sqrt{T})$ violation. The methods in  \cite{ShiEry_22, ZhouJi_22} assume Slater's condition and only consider soft violation $\sum_{t=1}^T g(x_t).$ 
The most closely related work is \cite{XuJiaJon_22}, which considered the problem of optimizing constrained black-box problems with the metric of cumulative violation. The work \cite{ LuPau_22} also proposed a penalty-based method to establish the bound on ``regret plus constraint violation", which unfortunately cannot provide individual bounds for regret and cumulative violation. The work \cite{XuJiaJon_22} requires solving an auxiliary constrained optimization problem at each round to keep the cumulative violation minimal. However, our paper designs an adaptive rectified framework to tackle the constraints, which leverages the penalty-based method to design the surrogate function and only needs to solve an unconstrained optimization problem at each round. 

{\noindent \bf Online Convex Optimization with Constraints} 
The online convex optimization with constraints has been widely studied in \cite{MahJinYan_12,SunDeyKap_17, NeeYu_17, ChaSal_19, CaoZhaPoo_21, SadRauFaz_20,YuaLam_18, YiLiYan_21a, YiLiYan_21b, GuoLiuWei_22}, where most of them consider constrained online convex optimization with soft constraint violation except \cite{YuaLam_18}, \cite{YiLiYan_21a} and \cite{GuoLiuWei_22} that study the metric of cumulative violation. \cite{YuaLam_18} developed an algorithm that achieves $O(\sqrt{T})$ regret and $O(T^{3/4})$ violation. \cite{YiLiYan_21a} improved the results to $O(\sqrt{T})$ regret and $O(T^{1/4})$ violation. \cite{GuoLiuWei_22} further improved the results to $O(\sqrt{T})$ regret and $O(1)$ violation. However, online convex optimization with constraints assumes the full information feedback, i.e., the complete form of objective and constraint functions, instead of the bandit feedback.  

\section{Problem Formulation}
We study a stochastic continuum-armed bandit with constraints, where the arms/decisions are in a continuous space $\mathcal X \in \mathbb R^d.$ The reward function $f: \mathcal X \to \mathbb R$ and constraint function $g: \mathcal X \to \mathbb R$ are continuous functions of the arms/decisions\footnote{We consider a single constraint for the ease of exposition and our results can be easily extended to the case with multiple constraints.}. Both $f$ and $g$ are black-box to the learner, and the learner acquires their knowledge sequentially.  
At each round $t \in [T]$, the learner makes decision $x_t \in\mathcal X$ and then observes the noisy reward and cost
\begin{align}
r_t = f(x_t) + \eta_t, ~~
c_t = g(x_t) + \varepsilon_t, \nonumber
\end{align}
where noise $\eta_t$ and $\varepsilon_t$ are random variables with zero-mean. Note $r_t$ and $c_t$ are bandit feedback because the leaner only observes the (noisy) version of $f(\cdot)$ and $g(\cdot)$ at $x_t.$ Since $f$ and $g$ are unknown apriori (possibly complicated and non-convex) and $\mathcal X$ is a continuous set with an infinity cardinality, 
it is infeasible in general to achieve the global optimal solution for arbitrary reward and constraint functions. We imposed the regularity assumption that $f(\cdot)$ and $g(\cdot)$ are within Reproducing Kernel Hilbert Space (RKHS). The assumption implies that a well-behaved continuous function can be represented with a properly chosen kernel function \cite{SriKraKak_09} and we can model reward and constraint functions via Gaussian processes as introduced below.\\
{\bf Gaussian process model for $f$ and $g$ functions} ~Gaussian process (GP) is a random process including a collection of random variables that follows a joint Gaussian distribution. Gaussian process $\text{GP}(\mu(x), k(x,x'))$ over $\mathcal X$ is specified by its mean $\mu(x)$ and covariance $k(x,x').$ For the reward function $f(x),$ we have $\text{GP}(\mu^f(x), k^f(x,x'))$ such that $\mu^f(x)=\mathbb E[f(x)]$ and $k^f(x,x') = \mathbb E[(f(x) - \mu^f(x))(f(x') - \mu^f(x'))].$  
Let $\mathcal A_t = \{x_1,\cdots,x_{t-1}\}$ be the collection of decisions and $\{r_1,\cdots,r_{t-1}\}$ be the collection of noisy feedback until round $t,$ respectively. The posterior distribution $\text{GP}(\mu^f_t(\cdot),k^f_t(\cdot,\cdot))$ updates at the beginning of round $t$
\begin{align}
    \mu^f_t(x) =& k^f_t(x)^T(V_t^f(\lambda))^{-1}r_{1:t} \label{gp:mean}\\
    k^f_t(x,x^{\prime}) =& k^f(x,x^{\prime}) - k^f_t(x)^T(V_t^f(\lambda))^{-1}k^f_t(x^{\prime}), \label{gp:kernel}\\
    \sigma_t^f(x) =& \sqrt{k_t^f(x,x)}, \label{gp:var}
\end{align}
where $K^f_t := [k^f(x,x^{\prime})]_{x,x^{\prime} \in \{x_1,\cdots,x_{t-1}\}},$ $V_t^f(\lambda):= K^f_t + \lambda I,$ $\lambda = 1+ 2/T$, $r_{1:t}=[r_1,\cdots,r_{t-1}],$ and $k^f_t(x) := [k^f(x_1,x),\cdots,k^f(x_{t-1},x)]^T.$ 
Similarly,
we define a GP model for the constraint function $g$ to be $\mathcal{GP}(\mu^g_t(x),k^g_t(x,x'))$ 
with the mean $\mu^g_t(x)$ and covariance $k^g_t(x,x').$ The model for $g$ updates the same as in \eqref{gp:mean}-\eqref{gp:var}
\begin{align}
    \mu^g_t(x) =& k^g_t(x)^T(V_t^f(\lambda))^{-1}c_{1:t} \label{gp:mean-g}\\
    k^g_t(x,x^{\prime}) =& k^g(x,x^{\prime}) - k^g_t(x)^T(V_t^g(\lambda))^{-1}k^g_t(x^{\prime}), \label{gp:kernel-g}\\
    \sigma_t^g(x) =& \sqrt{k_t^g(x,x)}, \label{gp:var-g}
\end{align}
where $K^g_t := [k^g(x,x^{\prime})]_{x,x^{\prime} \in \{x_1,\cdots,x_{t-1}\}},$  $V_t^g(\lambda):= K^f_t + \lambda I,$ $c_{1:t}=[c_1,\cdots,c_{t-1}],$ and $k^g_t(x) := [k^g(x_1,x),\cdots,k^g(x_{t-1},x)]^T.$ The kernel function is designed by choice and one popular kernel is the square exponential (SE) kernel $$k_{\text{SE}}(x,x') = e^{\frac{-\|x-x'\|^2}{2u^2}},$$
where $u > 0$ is a positive hyper-parameter. We consider the SE kernel function in this paper and use it in our experiments in Section \ref{sec: exp}.

Further, we define the information gain at round $t$ to be $\gamma^f_t := \max_{\mathcal A_t \in \mathcal{X}:|\mathcal A_t| = t-1} \frac{1}{2}\ln |I + \lambda^{-1} K^f_t|$ and $\gamma^g_t := \max_{\mathcal A_t \in \mathcal{X}:|\mathcal A_t| = t-1} \frac{1}{2}\ln |I + \lambda^{-1} K^g_t|$, which are important parameters in GP bandits. 
They depend on the choice of the kernel function and the domain $\mathcal X,$ and would play a key role in our following regret and violation analysis. For SE kernel function, we have $\gamma_t^f=O((\ln(t))^{d+1})$ and $\gamma_t^g=O((\ln(t))^{d+1})$ if $\mathcal X$ is compact and convex with dimension $d$. Next, we introduce the definition of regret and violation. 

{\bf Regret and cumulative constraint violation}~ 
Given the complete knowledge of $f$ and $g,$ we define the following offline optimization problem
\begin{align}
    \max_{x \in \mathcal X}& ~f(x) \label{def: obj}\\
    \hbox{s.t.}& ~~g(x) \leq 0. \label{def: cons}
\end{align}
Let $x^*$ be the global optimal solution to \eqref{def: obj}-\eqref{def: cons}. We define the regret and cumulative constraint violation 
\begin{align}
    \mathcal R(T) :=& \sum_{t=1}^T f(x^*) - \sum_{t=1}^T f(x_t), \label{def: regret}\\
    \mathcal V(T) :=& \sum_{t=1}^T g^+(x_t).  \label{def: violation}
\end{align}
The goal of the leaner is to develop algorithms to achieve sublinear regret and violation, i.e., $\lim_{T\to \infty} \mathcal R(T)/T = 0$ and $\lim_{T\to \infty} \mathcal V(T)/T = 0$ when $f$ and $g$ are modeled via Gaussian processes. 

\section{Rectified Pessimistic-Optimistic Learning Framework} \label{sec: GP-UCB}
In this section, we propose a general decision framework to tackle SCBwC with the metric of cumulative violation, called rectified pessimistic-optimistic learning framework (RPOL). The framework learns the reward function optimistically $\hat f_t(x)$ and the constraint function pessimistically $\check{g}_t(x)$ by a learning strategy $\mathcal M$ based on the model/parameters $(\Theta_{t}^f, \Theta_{t}^g).$ For example, the learning strategy could be the upper confidence bound learning of Gaussian process (GP-UCB), where $\Theta_{t}^{f}$ and $\Theta_{t}^{g}$ can include $(\mu_t^f, \sigma^f_t, k^f_t)$ and $(\mu_t^g, \sigma^g_t, k^g_t),$ respectively. 
By imposing the rectified operator on the constraint $\check g^{+}_t(x),$ RPOL chooses the best decision to maximize a rectified surrogate function $\hat f_t(x) - Q_t \check{g}_t^{+}(x)$ in \eqref{eq: decision}. After observing the noisy (possibly delayed) bandit feedback (reward and cost), we update the rectified penalty factor $Q_{t+1}$ and the model $(\Theta_{t+1}^f, \Theta_{t+1}^g),$ according to the learning strategy $\mathcal M.$ 
\vspace{0.1in}
\hrule
\vspace{0.1in}
\noindent{\bf RPOL Framework for SCBwC}
\vspace{0.1in}
\hrule
\vspace{0.1in}

\noindent {\bf Initialization:}   $Q_1 = 1$ and $\eta_t=\sqrt{t}.$ Model $\Theta_{1}^f$ and $\Theta_{1}^g.$

\noindent For $t=1,\cdots, T,$ 
\begin{itemize}
\item {\bf Pessimistic-optimistic learning:} estimate the reward function $\hat f_t(x)$ and the cost function $\check{g}_t(x)$ according to a learning strategy $\mathcal M$ with $(\Theta_{t}^f,\Theta_{t}^g).$  
\item {\bf Rectified penalty-based decision:} choose $x_{t}$ such that 
\begin{align}
x_{t} = \argmax_{x \in \mathcal X} ~ \hat f_{t}(x) - Q_t  \check{g}^{+}_t (x) \label{eq: decision} 
\end{align}
\item {\bf Feedback:} noisy reward $r_t(x_t)$ and cost $c_t(x_t).$ 
\item {\bf Rectified cumulative penalty update:} 
\begin{align}
  Q_{t+1} = \max\left(Q_t + c_t^+(x_t), \eta_t\right). \label{eq: rectified Q}
 \end{align}
\item {\bf Model update:} 
\begin{align}
    \Theta_{t+1}^f =& \mathcal M(\Theta_{t}^f, \{x_t, r_t, c_t\}),\\
    \Theta_{t+1}^g =& \mathcal M(\Theta_{t}^g, \{x_t, r_t, c_t\}). 
\end{align}
\end{itemize}
\vspace{0.1in}
\hrule
\vspace{0.1in}
We explain the main intuition behind the RPOL framework. 
The Lagrange function of the offline baseline problem in \eqref{def: obj}-\eqref{def: cons} is defined to be \[L(x, \lambda) :=  f(x) - \vartheta g(x),\] where $\vartheta$ is a dual variable related to the constraint in \eqref{def: cons}. Since the reward and cost functions are approximated via Gaussian Processes, we estimate $f(x)$ with $\hat f_t(x)$ optimistically and $g(x)$ with $\check{g}_t(x)$ pessimistically. We impose a rectified operator $\check{g}_t^{+}(x)$ to associate it with the hard violation ${g}_t^{+}(x)$ at round $t.$ Moreover, we approximate $\vartheta$ with a ``rectified'' penalty factor $Q_{t+1},$ where we first rectify the cost $c_t(x_t)$ with $c_t^+(x_t)$ and add it to $Q_{t}$ such that the penalty increases when the constraint violation occurs; and then we rectify $Q_{t+1}$ with a minimum penalty price $\eta_t.$ This design adaptively controls the penalty to prevent the aggressive decision for each round. 
The rectified decision in \eqref{eq: decision} and rectified penalty update in \eqref{eq: rectified Q} are the key to minimize the cumulative constraint $\sum_{t=1}^Tg_t^{+}(x).$

The ``rectified'' idea in this paper is motivated by \cite{GuoLiuWei_22} in online convex optimization with constraints. However, there exists a substantial difference due to the distinct feedback model: \cite{GuoLiuWei_22} observes the full-information feedback, imposes the rectifier on the previous constraint function, and introduces a smooth term to stabilize the learning process; this paper considers bandit feedback, learns the black-box functions (pessimistically and optimistically) directly and imposes a rectifier on the pessimistic estimator of constraint function. 
The ``rectified'' design also distinguishes our framework from the classical primal-dual approach in \cite{ZhouJi_22}. The work in \cite{ZhouJi_22} establishes the soft constraint violation (i.e., $\sum_{t=1}^T g(x_t)$) by studying the bound on the virtual queue/dual variable, which relies on the assumption of Slater's condition and the knowledge of slackness constant (the information is usually not available in practical applications). 
However, our framework establishes the cumulative violation (i.e., $\sum_{t=1}^T g^{+}(x_t)$) directly and does not require Slater's condition.

Before presenting theoretical results for the RPOL framework, we introduce the following two assumptions on reward function, constrained function, and noise. 
\begin{assumption}
Let $\|\cdot\|_k$ denote the RKHS norm associated with a kernel $k.$ For the reward function $f$, we assume that $\|f\|_{k^f} \leq B_f$ and $k^f(x,x) \leq 1$ for any $x \in \mathcal{X}$. For the constraint function $g$, we assume $\|g\|_{k^g} \leq B_g$ and $k^g(x,x) \leq 1$ for any $x \in \mathcal{X}$.
\label{assumption: function}
\end{assumption}

\begin{assumption}
The noise $\eta_t$ is i.i.d. $R_f$-sub-Gaussian and the noise $\varepsilon_t$ is i.i.d. $R_g$-sub-Gaussian.
\label{assumption: noise}
\end{assumption}

To establish a unified analysis method for SCBwC with the cumulative violation, we introduce a critical condition on the optimistic learning of reward function $\hat f$ and the pessimistic  learning of the constraint function $\check g,$ respectively.     

\begin{condition}
Let $\rho,$ $\{e^f_t(x)\},$ $\{e^g_t(x)\},$ be non-negative values. We have for any $x\in \mathcal X$ and all $t \in [T]$ such that
\begin{align*}
0 \leq \hat f_t(x) - \rho f(x) \leq e^f_t(x), \\
0 \leq \rho g(x) - \check{g}_t(x) \leq e^g_t(x), 
\end{align*}
hold with probability $1-p$ with $p \in (0,1).$ \label{cond}
\end{condition}

Intuitively, a good learning strategy $\mathcal M$ should satisfy Condition $1$ with small learning errors $e^f_t(x)$ and $e^g_t(x).$ These errors play  important roles in regret and cumulative violation in Theorem \ref{thm: main} as follows.
\begin{theorem}
Let Assumptions \ref{assumption: function} and \ref{assumption: noise} hold. Under Condition \ref{cond},
RPOL framework achieves the following regret and constraint violation
\begin{align*}
\mathcal R(T) \leq& \frac{1}{\rho}\sum_{t=1}^T e_t^f(x_t) , \\ 
\mathcal V(T) \leq& \frac{1}{\rho}\sum_{t=1}^T e_t^g(x_t)+ \sum_{t=1}^T e_t^f(x_t)  + 4\rho B_f \sqrt{T},
\end{align*} 
hold with the probability $1-p$ with $p \in [0,1].$
\label{thm: main}
\end{theorem}

\begin{remark}
RPOL framework is flexible to incorporate the classical learning strategies in unconstrained GP bandit learning (e.g., GP-UCB/LCB) and achieves strong performance guarantee on regret and cumulative violation for SCBwC in Theorem \ref{thm: main}. 
Moreover, RPOL framework can be readily combined with dedicated learning strategies for the variants of SCBwC and establish similar performance according to Theorem \ref{thm: main} as in the unconstrained counterparts.   
\end{remark}

In the following sections, we instantiate the learning strategies $\mathcal M$ in RPOL for SCBwC (and its variants), and establish the theoretical results according to Theorem \ref{thm: main}.

\section{Rectified Pessimistic-Optimistic Learning for SCBwC}\label{sec: classical}
In this section, we instantiate improved GP-UCB/LCB \cite{ChoSayGop_17} into RPOL framework for estimating $\hat f(x)$ and $\check g(x)$, and establish a strong performance on regret and violation according to Theorem \ref{thm: main}. 

{\noindent \bf GP-UCB/LCB}~ 
The optimistic estimator of $f(x)$ and the pessimistic estimator of $g(x)$ at round $t$ are defined by $$\hat f_t(x) = \mu^f_{t}(x) + \beta^f_t \sigma^f_{t}(x),~~ \check{g}_t(x) = \mu^g_{t}(x) - \beta^g_t \sigma^g_{t}(x),$$ 
which serves the upper confidence bound for the true $f(x)$ and the lower confidence bound for the true $g(x)$ by carefully choosing $\beta^f_t.$ 
We consider improved GP-UCB in \cite{ChoSayGop_17}. Let  $\beta^f_t=B_f+R_f\sqrt{2(\gamma^f_{t}+1+\ln{(2/p}))}$ and $\beta^g_t=B_g+R_g\sqrt{2(\gamma^g_{t}+1+\ln{(2/p}))}$ with $p \in (0,1).$ 
The models/parameters in GP-UCB/LCB, including $(\mu^f_t(x), \sigma_t^f(x), \mu^g_t(x), \sigma_t^g(x)),$ update according to \eqref{gp:mean}-\eqref{gp:var} and \eqref{gp:mean-g}-\eqref{gp:var-g}. We instantiate RPOL framework with GP-UCB/LCB into (RPOL-UCB) and present it as follows.  

\vspace{0.1in}
\hrule
\vspace{0.1in}
\noindent{\bf RPOL-UCB for SCBwC}
\vspace{0.1in}
\hrule
\vspace{0.1in}

\noindent {\bf Initialization:} $\mu^f_1(x) = \mu^g_1(x) = 0$, $\sigma^f_1(x) = \sigma^g_1(x) = 1$, $\forall x$, $Q_1 = 1$, $\eta_t=\sqrt{t},$ $\beta_t^f,$ and $\beta_t^g$.

\noindent For $t=1,\cdots, T,$ 
\begin{itemize}
\item {\bf Pessimistic-optimistic learning:} estimate the reward $\hat f_t(x)$ and the cost $\check{g}_t(x)$ with GP-UCB/LCB: 
\begin{align*}
    \hat f_t(x) = \mu^f_{t}(x) + \beta_t^f \sigma^f_{t}(x),~\check g_t(x) = \mu^g_t(x) - \beta_t^g  \sigma^g_{t}(x).
\end{align*}
\item {\bf Rectified penalty-based decision:} choose $x_{t}$ such that 
\begin{align*}
x_{t} = \argmax_{x \in \mathcal X} ~ \hat f_{t}(x) - Q_t  \check{g}^{+}_t (x) 
\end{align*}
\item {\bf Feedback:} noisy reward $r_t(x_t)$ and constraint $c_t(x_t).$ 
\item {\bf Rectified cumulative penalty update:} 
\begin{align}
  Q_{t+1} = \max\left(Q_t + c_t^+(x_t), \eta_t\right). \nonumber
 \end{align}
\item{\bf Posterior model update:} update $(\mu^f_{t+1}(x),\sigma^f_{t+1}(x))$ with $r_{1:t+1}$ according to \eqref{gp:mean}-\eqref{gp:var} and $(\mu^g_{t+1},\sigma^g_{t+1})$ with $c_{1:t+1}$ according to \eqref{gp:mean-g}-\eqref{gp:var-g}, respectively. 
\end{itemize}
\vspace{0.1in}
\hrule
\vspace{0.1in}

To analyze RPOL-UCB by Theorem \ref{thm: main}, we verify Condition \ref{cond} and quantify the cumulative errors for GP-UCB/LCB in Lemmas \ref{lem: GP-UCB} and \ref{lem: GP-UCB-error}, respectively. The detailed proof can be found in Appendix \ref{app: thm2}.
\begin{lemma}
Under Assumptions \ref{assumption: function} and \ref{assumption: noise},
the following inequalities hold for any $x \in \mathcal{X}$ and all $t \in [T]$ under RPOL-UCB
\begin{align*}
    0\leq \hat f_t(x) - f(x)  \leq 2\beta^f_{t} \sigma^f_{t}(x),\\
    0\leq g(x) - \check{g}_t(x) \leq 2\beta^g_{t} \sigma^g_{t}(x),
\end{align*}\label{lem: GP-UCB}
with probability at least $1 - p$ with $p\in (0,1).$
\end{lemma}

\begin{lemma}
Let $\{x_1,\cdots,x_T\}$ be the collection of decisions chosen by the algorithm. The cumulative standard deviation can be bounded as follows:
\begin{align*}
    &\sum_{t=1}^T \beta_t^f \sigma^f_{t}(x_t) \leq \beta_T^f\sqrt{4(T+2)\gamma^f_T}, \\ &\sum_{t=1}^T \beta_t^g\sigma^g_{t}(x_t) \leq \beta_T^g\sqrt{4(T+2)\gamma^g_T}.
\end{align*}
\label{lem: GP-UCB-error}
\end{lemma}
Based on Lemmas \ref{lem: GP-UCB} and \ref{lem: GP-UCB-error}, we invoke Theorem \ref{thm: main} to establish the regret and violation of RPOL-UCB in Theorem \ref{thm: GP-UCB}.
\begin{theorem}
RPOL-UCB achieves the following regret and constraint violation with a probability at least $1-p$:
\begin{align*}
\mathcal R(T) = O(\gamma_T\sqrt{T}), \\ 
\mathcal V(T) = O(\gamma_T\sqrt{T}),
\end{align*} 
where $\gamma_T = \max(\gamma^f_T,\gamma^g_T).$\label{thm: GP-UCB}
\end{theorem}
RPOL-UCB achieves a strictly stronger notation of cumulative violation compared to the soft violation in \cite{ShiEry_22, ZhouJi_22} and a similar performance compared to \cite{XuJiaJon_22} but with an efficient penalty approach.   
With the rectified design, RPOL quantifies the cumulative violation directly, which is different from the primal-dual optimization in \cite{ZhouJi_22} or the penalty-based technique in \cite{ShiEry_22, LuPau_22}. 

\section{RPOL for SCBwC with Delayed Feedback}\label{sec:delay}

In the previous section, we assume rewards feedback and costs/constraints feedback are available to the learner immediately. However, it might not happen in many real-world applications such as recommendation systems, clinical trials, and hyper-parameter tuning in machine learning, where the feedback is revealed to the learner after a random delay. Therefore, it motivates us to study SCBwC with stochastic delayed feedback.

At each round $t \in [T],$ the learner makes decision $x_t \in \mathcal X$ and observes the feedback
$$r_t = f(x_t) + \eta_t, ~~c_t = g(x_t) + \varepsilon_t$$ after stochastic delay $d^f_t$ and $d^g_t,$ respectively. We assume the delay $d^f_t$ and $d^g_t$ are independent and generated from an unknown distribution $\mathcal{D}$. 

To tackle the delayed feedback, we introduce the idea of censored feedback as in \cite{VerCarLat_20,VerDaiLow_22}. The delayed feedback is censored by indicator functions $\mathbb{I}\{d^f_s \leq \min (m,t-s)\}$ and $\mathbb{I}\{d^g_s \leq \min (m,t-s)\},$ which indicate if reward or cost at round $s$ are revealed by round $t$ and the delay is within $m$ rounds. We define the censored feedback at round $s$ by $\Tilde{r}_{s,t} := r_s \mathbb{I}\{d^f_s \leq \min (m,t-s)\}$ and $\Tilde{c}_{s,t} := c_s \mathbb{I}\{d^g_s \leq \min (m,t-s)\}$ and the sequence of censored feedback by $\Tilde{r}_{1:t} = [\Tilde{r}_{1,t-1}, \cdots, \Tilde{r}_{t-1,t-1}]^T;$ and $\Tilde{c}_{1:t}= [\Tilde{c}_{1,t-1}, \cdots, \Tilde{c}_{t-1,t-1}]^T.$ We further define $\rho^f_m = \mathbb{P}\{d^f_s \leq m\}$ and $\rho^g_m = \mathbb{P}\{d^g_s \leq m\},$ which denote the probabilities of
observing delayed reward feedback and cost feedback within $m$ rounds, respectively.

{\noindent \bf Censored GP-UCB/LCB} 
We utilize the censored feedback $\Tilde{r}_{1:t}$ (instead of $r_{1:t}$ in the previous section) when estimating the reward and constraint function $$\mu^f_t := k^f_t(x)^T(K^f_t + \lambda I)^{-1}\Tilde{r}_{1:t},~~\mu^g_t := k^f_t(x)^T(K^g_t + \lambda I)^{-1}\Tilde{c}_{1:t}.$$
The kernel matrix and variance update exactly the same as in \eqref{gp:kernel} and \eqref{gp:var}. Therefore, the optimistic and pessimistic estimators of $f(x)$ and $g(x)$ at round $t$ are $$\hat f_t(x) = \mu^f_{t}(x) + v^f_t \sigma^f_{t}(x),~~ \check g_t(x) = \mu^g_{t}(x) - v^g_t \sigma^g_{t}(x),$$ 
where $v^f_t = B_r\sum_{s=t-m}^{t-1}\sigma^f_{t}(x_s)+\beta^f_t$, $v^g_t = B_c\sum_{s=t-m}^{t-1}\sigma^g_{t}(x_s)+\beta^g_t$ with $B_r = B_f + R_f \sqrt{2\log T}$ and $B_c =  B_g + R_g \sqrt{2\log T}$ denoting bounds for observations $r_t$ and $c_t$ with the probability at least $1-2/T$ according Assumption \ref{assumption: noise}. Let $\beta^f_t = B_f + (R_f+B_r)\sqrt{2(\gamma^f_{t}+1+\ln{(4/p)})}$ and $\beta^g_t = B_g + (R_g+B_c)\sqrt{2(\gamma^g_{t}+1+\ln{(4/p)})},$ where $p \in (0,1).$ We instantiate RPOL framework with Censored GP-UCB/LCB  (RPOL-CensoredUCB). As the algorithm repeats most of the description of RPOL framework, we defer the complete description of RPOL-CensoredUCB to Appendix \ref{app: thm3}.

Similar to Section \ref{sec: classical}, we verify Condition \ref{cond} and quantify the cumulative errors for censored GP-UCB/LCB, and then invoke Theorem \ref{thm: main} to establish the following theorem. 
The detailed proof can be found in Appendix \ref{app: thm3}.

\begin{lemma}
Under Assumptions \ref{assumption: function}, \ref{assumption: noise},
the following inequalities hold for all $t \in [T]$ and $x \in \mathcal{X}$:
\begin{align*}
    0 \leq \hat f_t(x) -  \rho^f_m f(x) \leq 2v^f_t \sigma^f_{t}(x),\\
    0 \leq \hat g_t(x) -  \rho^g_m g(x) \leq 2v^g_t \sigma^g_{t}(x),
\end{align*}\label{lem: GP-UCB-delay}
\end{lemma}
\begin{lemma}
Let $\{x_1,\cdots,x_T\}$ be the collection of decisions selected by the algorithm. The cumulative standard deviation can be expressed in terms of the maximum information gain as:
\begin{align*}
    \sum_{t=1}^T v_t^f \sigma_t^f(x_t) \leq \beta_T^f \sqrt{4T\lambda \gamma^f_T} +  m B_r 4\lambda \gamma_T^f, \\
    \sum_{t=1}^T v_t^g \sigma_t^g(x_t) \leq \beta_T^g \sqrt{4T\lambda \gamma^g_T} +  m B_c 4\lambda \gamma_T^g.
\end{align*}
\label{lem: GP-UCB-delay-error}
\end{lemma}
Based on Lemmas \ref{lem: GP-UCB-delay} and \ref{lem: GP-UCB-delay-error}, we invoke Theorem \ref{thm: main} to establish
the regret and violation of RPOL with censored UCB/LCB (RPOL-CensoredUCB) in Theorem \ref{thm: delayGP}.
\begin{theorem}
RPOL with censored GP-UCB achieves the following regret and constraint violation with probability at least $1-p - 2/T$ with $p\in (0,1 - 2/T)$:
\begin{align*}
&\mathcal R(T) = O\left(\frac{\gamma_T}{\rho_m}(\sqrt{T} + m) + m\gamma_T)\right), \\ 
&\mathcal V(T) = O\left(\frac{\gamma_T}{\rho_m}(\sqrt{T} + m) + m\gamma_T\right),
\end{align*} 
where $\gamma_T = \max(\gamma^f_T,\gamma^g_T)$ and $\rho_m = \min(\rho_m^f,\rho_m^g)$.
\label{thm: delayGP}
\end{theorem}
Theorem \ref{thm: delayGP} shows that RPOL-CensoredUCB achieves sub-linear bounds for the regret and violation simultaneously in SCBwC with delayed feedback. The result matches the regret bound for unconstrained counterparts with delayed feedback in \cite{VerDaiLow_22}.

\section{RPOL for SCBwC under Non-stationary Environment}\label{sec:non-stationary}
The previous sections assume the reward function $f$ and constraint function $g$ are time-invariant.  
However, both functions $f$ and $g$ might change as times in many real-world applications. For example, in energy-efficient job scheduling in data centers, the arrival rates of the incoming jobs and energy prices fluctuate from time to time. To capture the non-stationary environment, we introduce the definition of variation budget
$$
    P_T = \max(\sum_{t=1}^T\|f_{t+1}-f_t\|_{k^f},\sum_{t=1}^T\|g_{t+1}-g_t\|_{k^g}). \nonumber
$$
Such a variation budget model is common in non-stationary bandit learning \cite{ZhouShr_21,DenZhoGho_22} and non-stationary online convex optimization \cite{Zin_03,HalWil_13,ZhaLuZho_18}.

The feedback model is similar to that in Section \ref{sec: classical}. At each round $t \in [T]$, the learner makes decision $x_t$ and then observes a bandit reward feedback $r_t = f_t(x_t) + \eta_t$ and a bandit constraint feedback $c_t = g_t(x_t) + \varepsilon_t$, where $f_t$ and $g_t$ are time-varying and block-box functions. For the non-stationary setting, we define the following dynamic baseline 
\begin{align*}
    \mathcal R(T) := \sum_{t=1}^T f_t(x_t^*) - \sum_{t=1}^T f_t(x_t), ~~
    \mathcal V(T) := \sum_{t=1}^T g_t^+(x_t), 
\end{align*}
where $x^*_t$ denotes the solution to 
\begin{align*}
    \max_{x \in \mathcal{X}} f_t(x),~s.t.~g_t(x)\leq 0.
\end{align*}
{\bf Sliding Window GP-UCB} To address the non-stationary challenges, we consider sliding window GP-UCB/LCB (SW-UCB/LCB), which has been shown to guarantee a sub-linear dynamic regret bound for the unconstrained GP bandits \cite{ZhouShr_21}. The sliding window approach abandons outdated data and utilizes the latest observations within a window with size $W$. The SW-UCB/LCB estimators are defined as follows
\begin{align*}
    \hat f_t(x) = \mu^f_{t}(x) + \beta^f_t \sigma^f_{t}(x)+ \Gamma_t^f,~~
    \check g_t(x) = \mu^g_{t}(x) - \beta^g_t \sigma^g_{t}(x)- \Gamma_t^g ,
\end{align*}
where $\Gamma^f_t = C_f \sum_{s = t_0}^{t-1}\|f_s - f_{s+1}\|_{k^f}$, $\Gamma^g_t = C_g \sum_{s = t_0}^{t-1}\|g_s - g_{s+1}\|_{k^g}$ with $C_f = \frac{1}{\lambda} \sqrt{2W(1+\lambda)\gamma^f_T}$ and $C_g = \frac{1}{\lambda} \sqrt{2W(1+\lambda)\gamma^g_T}$. Let $\beta^f_t = B_f + \frac{1}{\sqrt{\lambda}} R_f \sqrt{2\gamma^f_{t-t_0} + 2 \ln (2T/p)}$ and $\beta^g_t = B_g + \frac{1}{\sqrt{\lambda}} R_g \sqrt{2\gamma^g_{t-t_0} + 2 \ln (2T/p)},$ where $t_0$ denotes the beginning of the sliding window. 
We instantiate RPOL with sliding window GP-UCB/LCB (ROPL-SWUCB). As the algorithm repeats most of the description of RPOL framework, we also defer the complete description of RPOL-SWUCB to Appendix \ref{app: thm4}.

Similar to Sections \ref{sec: classical} and \ref{sec:delay}, we verify Condition \ref{cond} and quantify the cumulative errors for SW-UCB/LCB, and then invoke Theorem \ref{thm: main} to establish the following theorem for SCBwC under non-stationary environment. 
The detailed proof can be found in Appendix \ref{app: thm4}.

\begin{lemma}
Under Assumptions \ref{assumption: function}, \ref{assumption: noise},
the following inequalities hold for all $t \in [T]$ and $x \in \mathcal{X}$:
\begin{align*}
     &0 \leq \hat f_{t}(x) - f_t(x) \leq 2\Gamma^f_t + 2\beta^f_t \sigma^f_{t}(x),\\
     &0 \leq g_{t}(x) - \check g_t(x) \leq 2\Gamma^g_t + 2\beta^g_t \sigma^g_{t}(x),
\end{align*}
where $\Gamma^f_t = \frac{1}{\lambda} \sqrt{2W(1+\lambda)\gamma^f_T} \sum_{s = t_0}^{t-1}\|f_s - f_{s+1}\|_{k^f}$ and $\Gamma^g_t = \frac{1}{\lambda} \sqrt{2W(1+\lambda)\gamma^g_T} \sum_{s = t_0}^{t-1}\|g_s - g_{s+1}\|_{k^g}$.
\label{lem: GP-UCB-SW}
\end{lemma}

\begin{lemma}
Let $\{x_1,\cdots,x_T\}$ be the collection of decisions selected by the algorithm. The cumulative standard deviation can be expressed in terms of the maximum information gain as:
\begin{align*}
    &\sum_{t=1}^T \Gamma^f_t\leq C_f W P_T ,~  \sum_{t=1}^T \beta^f_t \sigma^f_{t}(x_t)\leq \beta_T^f T\sqrt{\frac{4\lambda \gamma_T^f}{W}}, \\
  &\sum_{t=1}^T \Gamma^f_t\leq C_g W P_T ,~  \sum_{t=1}^T \beta^g_t \sigma^g_{t}(x_t)\leq \beta_T^g T\sqrt{\frac{4\lambda \gamma_T^g}{W}},
\end{align*}
where we define the coefficients $C_f = \frac{1}{\lambda} \sqrt{2W(1+\lambda)\gamma^f_T}$ and $C_g = \frac{1}{\lambda} \sqrt{2W(1+\lambda)\gamma^g_T}$.
\label{lem: GP-UCB-SW-error}
\end{lemma}
Based on Lemmas \ref{lem: GP-UCB-SW} and \ref{lem: GP-UCB-SW-error}, we invoke Theorem \ref{thm: main} to establish
the regret and violation of in Theorem \ref{thm: dynamicGP} for RPOL-SWUCB.

\begin{theorem}
Let the window size $W = \gamma_T^{1/4}(T/P_T)^{1/2}$ and RPOL-SWUCB achieves the following regret and constraint violation with the probability at least $1-p$
\begin{align*}
\mathcal R(T) =O(\gamma_T^{7/8} P_T^{1/4}T^{3/4}), \\ \mathcal V(T) =O(\gamma_T^{7/8} P_T^{1/4}T^{3/4}). 
\end{align*}
\label{thm: dynamicGP}
\end{theorem}

\begin{remark}
The choice of window size $W$ depends on the knowledge of the path-length $P_T$. It is common to assume the knowledge of $P_T$ or its upper bound is available in the literature of non-stationary bandits \cite{CheSimZhu_19,CheSimZhu_22,ZhouShr_21,ZhaZhaJia_20,LiuJiaLi_22}. For the practical applications where the knowledge of $P_T$ is hard to be estimated, one can utilize a general reduction technique recently developed by \cite{WeiLuo_21} to achieve similar regret and violation bounds without the prior knowledge of $P_T.$ 
\end{remark}

\section{Experiments} \label{sec: exp}

In this section, we test the performance of RPOL framework with numerical experiments and compare our algorithms with the existing baselines in the following three settings. We plot the average regret and violation $\mathcal R(t)/t$ and $\mathcal V(t)/t.$

\begin{figure}[H]
\centering
\begin{subfigure}{0.43\textwidth}
    \includegraphics[width=\textwidth]{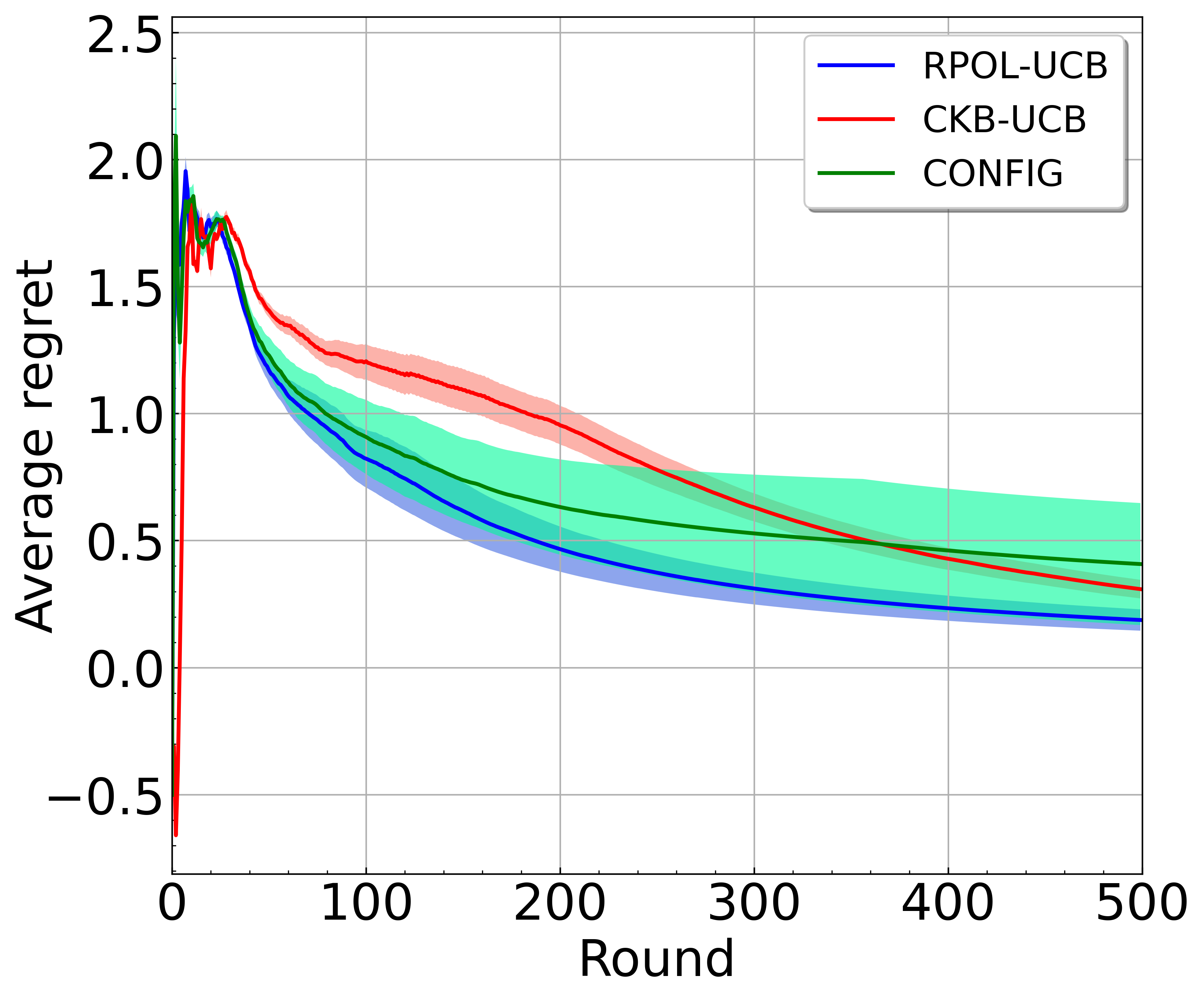}
    \caption{Average Regret}
    \label{fig:env1-regret}
\end{subfigure}
\hfill
\begin{subfigure}{0.43\textwidth}
    \includegraphics[width=\textwidth]{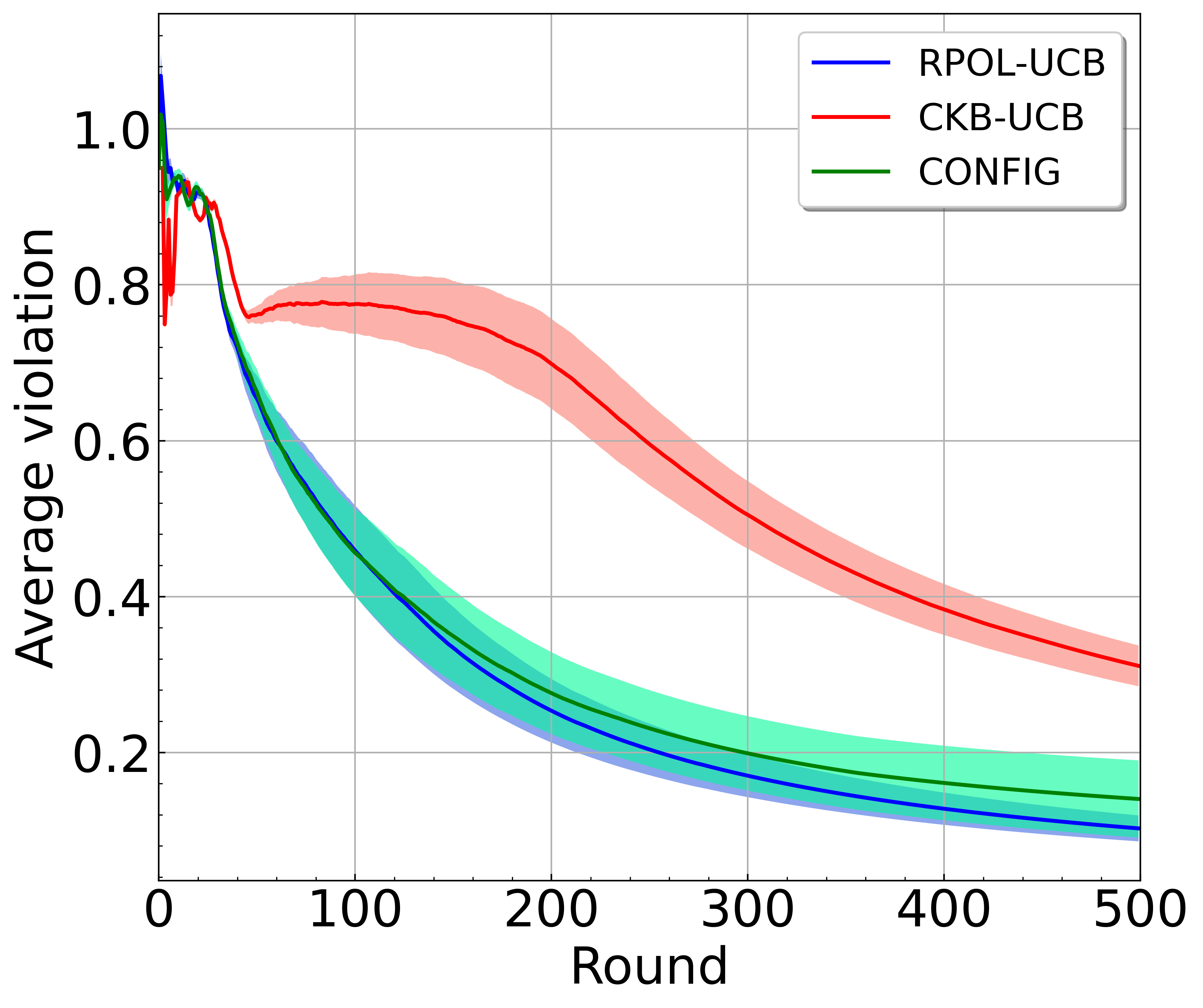}
    \caption{Average Violation}
    \label{fig:env1-violation}
\end{subfigure}
\caption{Regret and Cumulative Violation in SCBwC}  
\label{fig:traditional}
\end{figure}

\vspace{5 pt}
{\noindent \bf Classical SCBwC}
We consider the reward function $f(x) = -\sin{x(1)} - x(2)$ and the constraint function $g(x) = \sin{x(1)}\sin{x(2)} + 0.95,$ where $x \in [0,6]^2.$ The constraint set $\{x~|~g(x) \leq 0\}$ indicates a strict region and makes the problem challenging.  The observations are corrupted with Gaussian noise sampled from $\mathcal{N}(0,0.05),$ respectively. We test RPOL-UCB and consider the baselines: CKB-UCB in \cite{ZhouJi_22} and CONFIG in  \cite{XuJiaJon_22}. From Figure \ref{fig:env1-regret} and \ref{fig:env1-violation}, we show RPOL-UCB achieves the best performance w.r.t. both regret and cumulative violation in SCBwC, where it converges to a low cumulative violation in a faster rate. The results in Figure \ref{fig:env1-regret} and \ref{fig:env1-violation} justify that our rectified design can balance the regret and cumulative violation efficiently and safely, and it is superior to handling the strict cumulative violation. 

\begin{figure}[H]
\centering
\begin{subfigure}{0.43\textwidth}
    \includegraphics[width=\textwidth]{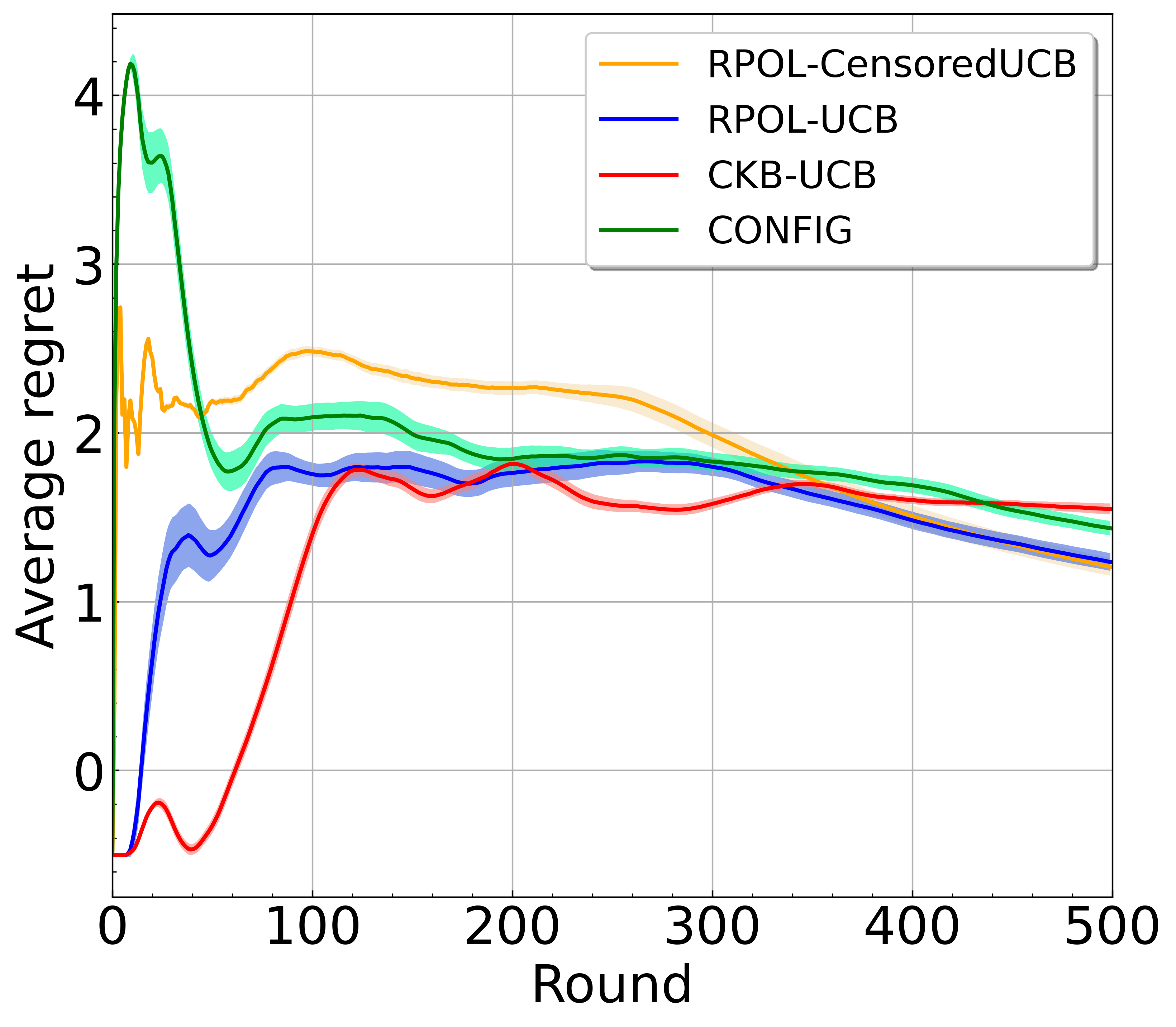}
    \caption{Average Regret}
    \label{fig:delay-env1-regret}
\end{subfigure}
\hfill
\begin{subfigure}{0.43\textwidth}
    \includegraphics[width=\textwidth]{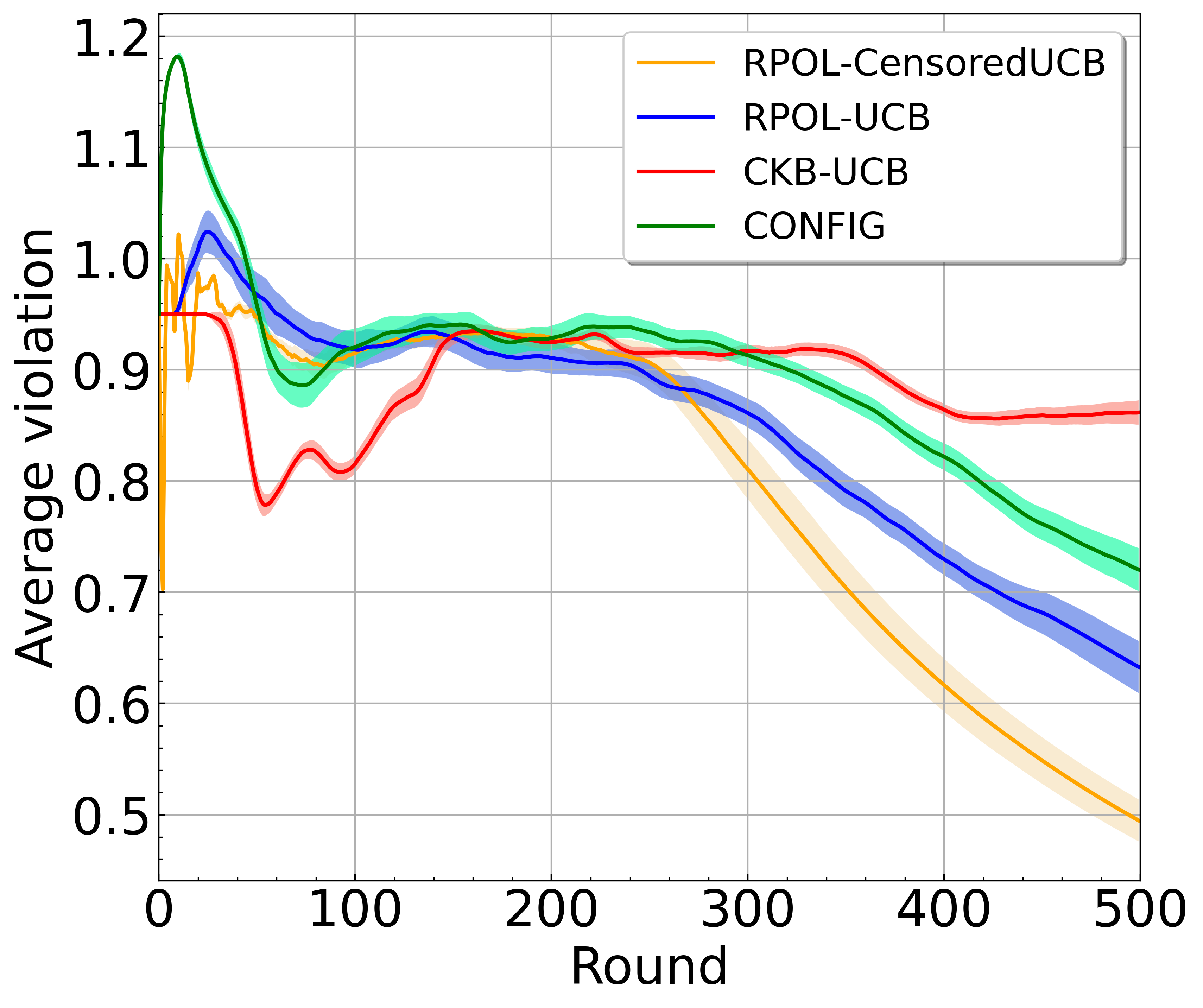}
    \caption{Average Violation}
    \label{fig:delay-env1-violation}
\end{subfigure}
\caption{Regret and Cumulative Violation in SCBwC with Delayed Feedback}  
\label{fig:delay}
\end{figure}

\vspace{5 pt}
{\noindent \bf SCBwC with delayed feedback} We consider the stochastic delayed feedback based on the first experiment, where the delay of $d_t^f$ and $d_t^g$ at round $t$ are sampled from a Poisson distribution with mean $15,$ respectively. We test RPOL-CensoredUCB and consider RPOL-UCB, CKB-UCB, and CONFIG as the baseline algorithms. From \ref{fig:delay-env1-regret} and \ref{fig:delay-env1-violation}, RPOL-CensoredUCB also outperforms all existing baselines. These results indicate RPOL framework can establish a strong performance guarantee even with stochastic delayed feedback. 

\begin{figure}[H]
\centering
\begin{subfigure}{0.43\textwidth}
    \includegraphics[width=\textwidth]{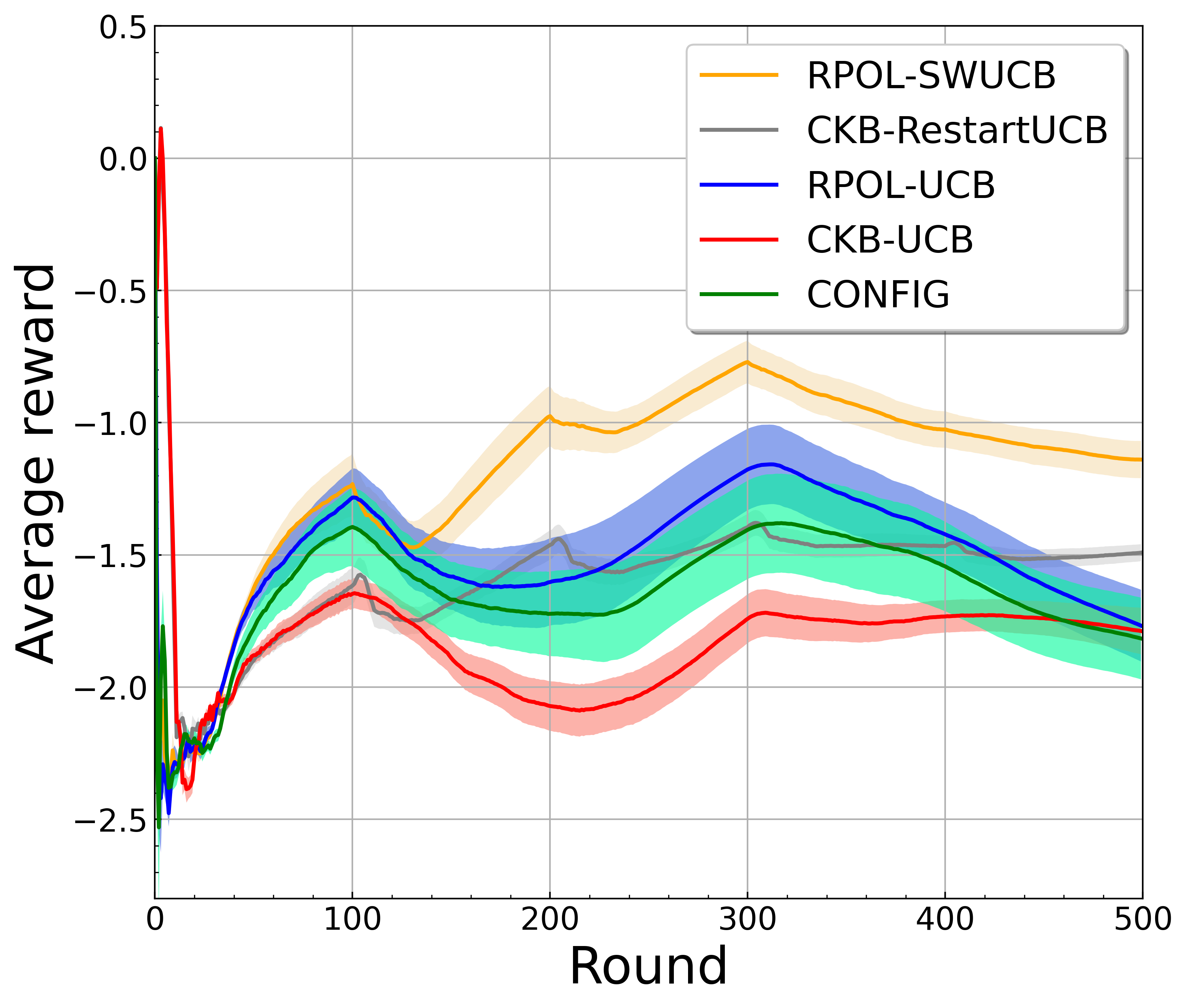}
    \caption{Average Reward}
    \label{fig:non-stationary-env1-regret}
\end{subfigure}
\hfill
\begin{subfigure}{0.43\textwidth}
    \includegraphics[width=\textwidth]{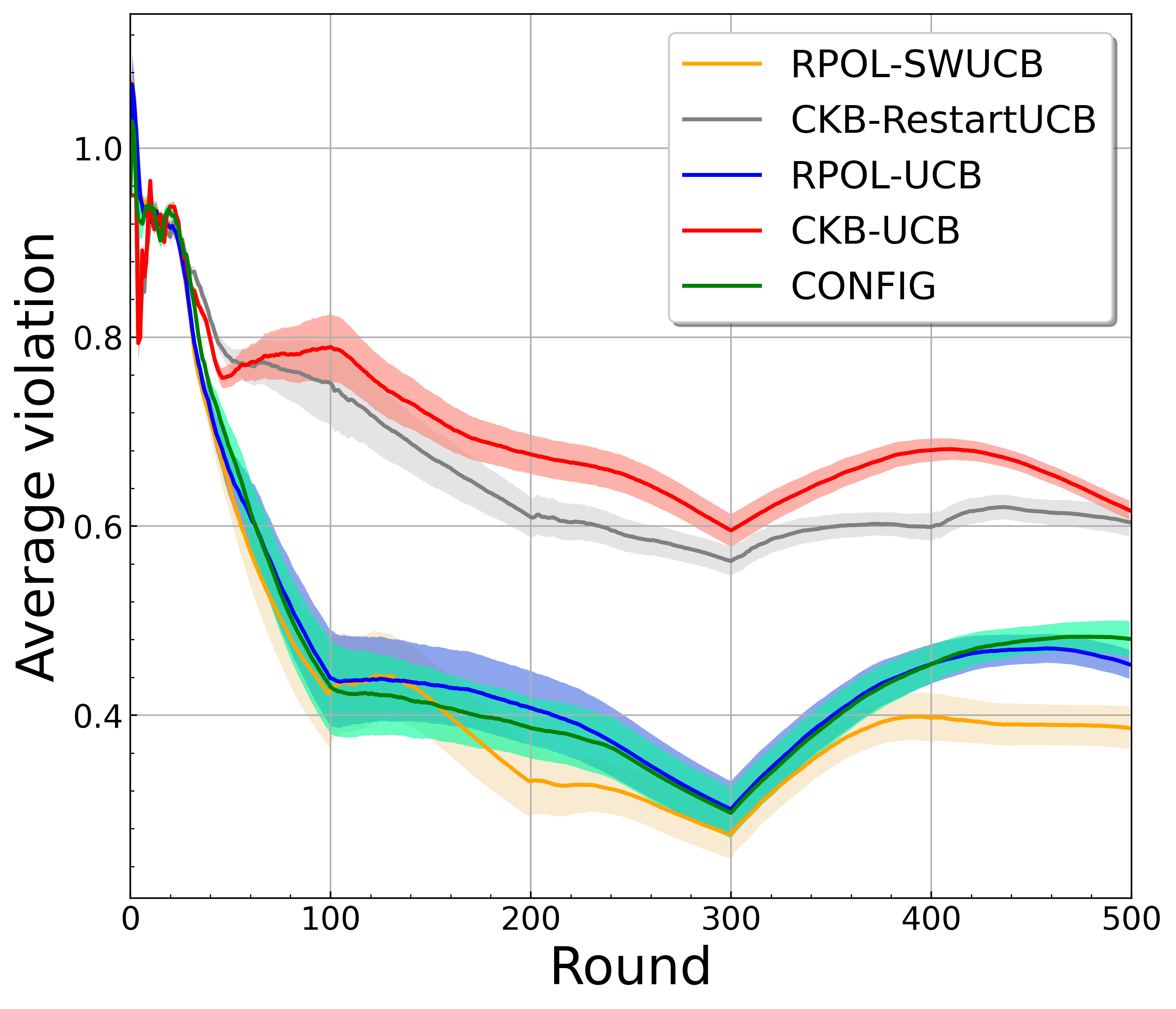}
    \caption{Average Violation}
    \label{fig:non-stationary-env1-violation}
\end{subfigure}
\caption{Regret and Cumulative Violation in SCBwC under Non-stationary Environment}  
\label{fig:non-stationary}
\end{figure}

\vspace{5 pt}
{\noindent \bf SCBwC under non-stationary environment} We consider the non-stationarity based on the first experiment, where the reward function and constraint function vary at $100$ and $300$ round. Specifically, we set $f_{1:100} = -\sin x(1) - x(2)$; $g_{1:100} = \sin x(1) \sin x(2) + 0.95$; $f_{101:300} = -\sin (x(1)-5) - x(2)$; $g_{101:300} = \sin x(1) \sin (x(2)+5) + 0.5$; $f_{301:500} = -\sin (x(1)+4) - x(2)$; $g_{301:500} = \sin (x(1)+5) \sin (x(2)) + 0.95$. We test RPOL-SWUCB and consider CKB-UCB, CONFIG, and CKB-RestartUCB in \cite{DenZhoGho_22}. From \ref{fig:non-stationary-env1-regret} and \ref{fig:non-stationary-env1-violation}, we again observe that RPOL-SWUCB has the best performance. It demonstrates that our RPOL framework is flexible and efficient in the non-stationary environment.

\section{Conclusion}
In this paper, we study stochastic
continuum-armed bandit with constraints with the cumulative constraint violation. We propose the rectified pessimistic-optimistic learning framework and show it is flexible to be applied into stochastic continuum-armed bandit with constraints and its variants by utilizing the dedicated exploration techniques. We develop unified analysis techniques to show our framework is efficient in achieving sublinear regret and cumulative violation. Our theoretical and experimental results justify the superior of the proposed framework.

\bibliographystyle{plain}
\bibliography{ref}

\newpage
\onecolumn
\appendix

\section{Proof of Theorem \ref{thm: main}}
To prove Theorem \ref{thm: main}, we first introduce a key ``self-bounding property" to establish an upper bound on ``regret + cumulative violation'', motivated by \cite{GuoLiuWei_22}.\\
{\bf Self-bounding property:}  From the decision choice of $x_t$ in \eqref{eq: decision}, we have for any $x \in \mathcal{X}$ such that $$\hat f_{t}(x)-Q_t\check{g}^{+}_t (x) \leq  \hat f_{t}(x_t) - Q_t  \check{g}^{+}_t (x_t), ~\forall t\in[T].$$ Let $x = x^*$ and add $\rho(f(x^*)-f(x_t))$ to both sides of the inequality above
\begin{align*}
    &\rho(f(x^*)-f(x_t)) + \hat f_{t}(x^*)-Q_t\check{g}^{+}_t (x^*) \\
    \leq& \rho(f(x^*)-f(x_t)) + \hat f_{t}(x_t) - Q_t  \check{g}^{+}_t (x_t).
\end{align*}
From Condition \ref{cond}, we have $$\check g_t(x^*) \leq g_t(x^*) \leq 0, ~\forall t\in[T].$$ hold with a high probability at least $1-p.$ Since $Q_t \geq 1$ according to its definition, we rearrange the inequality above and have for any $t\in [T]$
\begin{align}
    &\rho(f(x^*)-f(x_t)) + Q_t  \check{g}^{+}_t (x_t) \nonumber\\ 
    \leq& \rho f(x^*) - \hat f_{t}(x^*)  + \hat f_{t}(x_t) - \rho f(x_t). \label{eq: self-bound}
\end{align}
Based on the ``self-bounding property" in \eqref{eq: self-bound}, we establish the regret and violation in Theorem \ref{thm: main}.

{\noindent \bf Regret bound:} Since $Q_{t} \geq 1,$ we have $Q_{t} \check{g}^{+}_t(x_t) \geq 0.$ The inequality \eqref{eq: self-bound} implies  
\begin{align*}
    &\rho (f(x^*) - f(x_t)) \\
    \leq& \rho f(x^*) - \hat f_{t}(x^*)  + \hat f_{t}(x_t) - \rho f(x_t).
\end{align*}
From Condition \ref{cond}, we have $f(x^*) - \hat f_{t}(x^*) \leq 0$ and $\hat f_{t}(x_t) - f(x_t) \leq e_t^f(x_t)$ for all $t \in [T]$ with the probability at least $1-p.$ We have 
\begin{align*}
        \mathcal R(T):=&\frac{1}{\rho} \sum_{t=1}^T \rho (f(x^*) - f(x_t)) \\
                \leq& \frac{1}{\rho}\sum_{t=1}^T (\hat f_{t}(x_t) - \rho f(x_t)) +\frac{1}{\rho}\sum_{t=1}^T(\rho f(x^*) - \hat f_{t}(x^*)) \\
                \leq& \frac{1}{\rho}\sum_{t=1}^T e_t^f(x_t) 
\end{align*}
holds with the probability at least $1-p.$\\
{\bf Violation bound:} 
We first establish the upper bound of $\sum_{t=1}^T \check{g}^{+}_t(x_t)$ and then connect it with $\sum_{t=1}^T g^{+}_t(x_t).$ 

Rearrange inequality \eqref{eq: self-bound} and we have:
\begin{align*}
    Q_{t} \check{g}^{+}_t(x_t) \leq& \rho f(x^*) - \hat f_{t}(x^*)  + \hat f_{t}(x_t) - \rho f(x_t)\\ &+  \rho ( f(x_t)-f(x^*)),
\end{align*}
which, conjunction with Condition \ref{cond}, implies 
\begin{align*}
    \check{g}^{+}_t(x_t) 
    \leq \frac{e_t^f(x_t) + \rho (f(x_t)-f(x^*))}{Q_{t}}.
\end{align*}
Since $Q_{t} \geq \eta_t = \sqrt{t}$ according to the definition, we have
\begin{align}
    \sum_{t=1}^T \check{g}^{+}_t(x_t) \leq &\sum_{t=1}^T\frac{e_t^f(x_t) }{\sqrt{t}}  + \rho \sum_{t=1}^T \frac{ f(x_t)-f(x^*)}{\sqrt{t}} \nonumber\\
    \leq &\sum_{t=1}^T e_t^f(x_t) + 4\rho B_f\sqrt{T}, \label{eq: g check}
\end{align}
where the last inequality holds because (i) $t \geq 1;$ (ii) $f$ is bounded by $B_f$ in Assumption \ref{assumption: function}; and (iii) $\sum_{t=1}^T \frac{1}{\sqrt{t}} \leq \int_{1}^T \frac{1}{\sqrt{t}} dt \leq 2 \sqrt{T}$.

Next, we establish violation based on the relationship between $\check{g}^{+}_t(x_t)$ and ${g}^{+}(x_t)$ in Condition \ref{cond} as follows
\begin{align*}
    \mathcal V(T) :=& \frac{1}{\rho}\sum_{t=1}^T \rho g^+(x_t) \\
    =&  \frac{1}{\rho}\sum_{t=1}^T (\rho g(x_t) - \check{g}_t(x_t) + \check{g}_t(x_t))^+ \\
    \leq&  \frac{1}{\rho} \sum_{t=1}^T (\rho g(x_t) - \check{g}_t(x_t))^+ + \sum_{t=1}^T \check{g}_t^+(x_t) \\
    \leq&  \frac{1}{\rho} \sum_{t=1}^T  e_t^g(x_t)  + \sum_{t=1}^T \check{g}_t^+(x_t) \\
    \leq&  \sum_{t=1}^T e_t^f(x_t) +  4\rho B_f\sqrt{T} + \frac{1}{\rho} \sum_{t=1}^T e_t^g(x_t) 
\end{align*}
where the first inequality holds because $(a+b)^{+} \leq (a)^{+} + (b)^{+};$ the second inequality holds because of Condition \ref{cond}; the last inequality holds by \eqref{eq: g check}.

\section{Proof of Theorem \ref{thm: GP-UCB}} \label{app: thm2}
In this section, we prove regret bound and violation bound for RPOL with GP-UCB (RPOL-UCB) for SCBwC by Theorem \ref{thm: main}. To invoke Theorem \ref{thm: main}, we need to verify Condition \ref{cond} by proving Lemmas \ref{lem: GP-UCB} and \ref{lem: GP-UCB-error}. 

\subsection{Proof of Lemma \ref{lem: GP-UCB}} \label{app: lemma_2}
Lemma \ref{lem: GP-UCB} establishes the confidence bounds for estimators $\hat{f}_t(\cdot)$ and $\check{g}_t(\cdot)$. We first prove for the reward function $f$ and the analysis for the constraint function $g$ follows the same steps.

According to $\hat{f}_t(x) = \mu^f_t(x) + \beta^f_t \sigma^f_t(x)$, we have
\begin{align*}
     \hat{f}_t(x) - f(x) = \mu_t^f(x) - f(x)+ \beta_t^f\sigma_t^f(x).
\end{align*}
Recall that the reward function $f$ lies in RKHS. For convenience, we define $\varphi(x) = k^f(x,\cdot)$ instead of $\varphi^f(x)$, it implies $f(x) = \langle f, k^f(x,\cdot)\rangle_{k^f} = \langle f, \varphi(x)\rangle_{k^f} := f^T\varphi(x).$ 
Further define the RKHS norm $\|g\|_{k^f}$ as $\sqrt{g^T g}$, $\Phi_t = [\varphi(x_1)^T, \ldots,\varphi(x_{t-1})^T]^T$, then kenerl matrix $K^f_t = \Phi_t \Phi_t^T$, $k^f_t(x) = \Phi_t \varphi(x)$ for all $x \in \mathcal{X}$ and $f_{1:t} = \Phi_t f$. Since $\mu^f_t(x) = k^f_t(x)^T(K_t^f + \lambda I)^{-1}r_{1:t}$, and $r_{1:t} = f_{1:t} + \varepsilon_{1:t}$, we have
\begin{align}
    \mu_t^f(x) - f(x) = k^f_t(x)^T(K_t^f + \lambda I)^{-1}f_{1:t} - f(x) + k^f_t(x)^T(K_t^f + \lambda I)^{-1}\varepsilon_{1:t}, \label{diff terms}
\end{align}
From Theorem 2 in \cite{ChoSayGop_17}, we derive the difference term in \eqref{diff terms} as follows
\begin{align*}
    |k^f_t(x)^T(K_t^f + \lambda I)^{-1}f_{1:t} - f(x)| &= |\varphi(x)^T \Phi_t^T(\Phi_t\Phi_t^T + \lambda I)^{-1}\Phi_t f - \varphi(x)^T f| \\
    &=|\varphi(x)^T (\Phi_t^T\Phi_t + \lambda I)^{-1}\Phi_t^T\Phi_t f - \varphi(x)^T f| \\
    &=|\lambda \varphi^T(\Phi_t^T\Phi_t + \lambda I)^{-1}f| \\
    &\leq \| \lambda (\Phi_t^T\Phi_t + \lambda I)^{-1} \varphi(x)\|_{k^f} \| f \|_{k^f} \\
    &= \| f \|_{k^f} \sqrt{\lambda\varphi(x)^T (\Phi_t^T \Phi_t + \lambda I)^{-1} \lambda I (\Phi_t^T \Phi_t + \lambda I)^{-1} \varphi(x)} \\
    &\leq B_f \sqrt{\lambda \varphi(x)^T(\Phi_t^T\Phi_t + \lambda I)^{-1}(\Phi_t^T\Phi_t + \lambda I)(\Phi_t^T\Phi_t + \lambda I)^{-1}\varphi(x)} \\
    &= B_f \sigma^f_t(x),
\end{align*} 
where the second equality comes from the fact that $\Phi_t^T(\Phi_t\Phi_t^T + \lambda I)^{-1} = (\Phi_t^T\Phi_t + \lambda I)^{-1}\Phi_t^T$ and the third equality comes from $\varphi(x) = \Phi_t^T (\Phi_t\Phi_t^T + \lambda I)^{-1} k_t^f(x) + \lambda (\Phi_t^T\Phi_t + \lambda I)^{-1} \varphi(x),$ and this implies $\lambda \varphi(x)^T(\Phi_t^T\Phi_t + \lambda I)^{-1}\varphi(x) = k^f(x,x) - k^f_t(x)^T(K_t^f + \lambda I)^{-1} k_t^f(x) = (\sigma^f_t(x))^2$ and prove the last equality.

For the second term in \eqref{diff terms}, we have
\begin{align*}
   |k^f_t(x)^T(K_t^f + \lambda I)^{-1}\varepsilon_{1:t}| &= |\varphi(x)^T\Phi_t^T(\Phi_t\Phi_t^T + \lambda I)^{-1}\varepsilon_{1:t}|\\
   &= |\varphi(x)^T(\Phi_t\Phi_t^T + \lambda I)^{-1}\Phi_t^T\varepsilon_{1:t}| \\
   &\leq \|(\Phi_t\Phi_t^T + \lambda I)^{-1/2}\varphi(x)\|_{k^f} \|(\Phi_t\Phi_t^T + \lambda I)^{-1/2}\Phi_t^T \varepsilon_{1:t}\|_{k^f}\\
   &= \sqrt{\varphi(x)^T(\Phi_t^T \Phi_t + \lambda I)^{-1} \varphi(x)} \sqrt{(\Phi_t^T \varepsilon_{1:t})^T(\Phi_t^T\Phi_t + \lambda I)^{-1}\Phi_t^T \varepsilon_{1:t}} \\
   &= \lambda^{-1/2} \sigma^f_t(x) \sqrt{\varepsilon_{1:t}^T K_t^f(K_t^f + \lambda I)^{-1}\varepsilon_{1:t}} \\
   &\leq \lambda^{-1/2} \sigma^f_t(x) \sqrt{\varepsilon_{1:t}^T (K_t^f + \eta I)(K_t^f + \lambda I)^{-1}\varepsilon_{1:t}} \\
   &\leq \sigma^f_t(x) \sqrt{\varepsilon_{1:t}^T ((K^f_t + (2/T) I)^{-1} + I)^{-1}\varepsilon_{1:t}} \\
   &\leq \sigma^f_t(x) \| \varepsilon_{1:t} \|_{((K^f_t + (2/T) I)^{-1} + I)^{-1}},
\end{align*}
where the first inequality comes from $\Phi_t^T(\Phi_t\Phi_t^T + \lambda I)^{-1} = (\Phi_t^T\Phi_t + \lambda I)^{-1}\Phi_t^T$ and $\lambda \varphi(x)^T(\Phi_t^T\Phi_t + \lambda I)^{-1}\varphi(x) = k^f(x,x) - k^f_t(x)^T(K_t^f + \lambda I)^{-1} k_t^f(x)$. According to Theorem 1 in \cite{ChoSayGop_17},
we have for any $x \in \mathcal{X}$ and for all $t \in [T]$
\begin{align*}
    |k^f_t(x)^T(K_t^f + \lambda I)^{-1}\varepsilon_{1:t}| \leq  R_f\sqrt{2(\gamma^f_t + 1 + \ln(2/p))} \sigma^f_t(x)
\end{align*}
holds with the probability at least $1 - p/2.$ Recall the definition of $\beta^f_t,$ we have for any $x \in \mathcal{X}$ and for all $t \in [T]$
\begin{align*}
    -\beta^f_t\sigma^f_t(x) + \beta^f_t\sigma^f_t(x) \leq \hat{f}_t(x) - f(x) \leq 
    \beta^f_t\sigma^f_t(x) + \beta^f_t\sigma^f_t(x).
\end{align*} holds with the probability at least $1 - p/2.$ 
Follow the same steps for the constraint function of $g,$ we complete the proof.

\subsection{Proof of Lemma \ref{lem: GP-UCB-error}}
We prove the first inequality on reward function $f$ in Lemma \ref{lem: GP-UCB-error} and  the second inequality on $g$ holds by following the same steps. By Cauchy-Schwartz inequality, we have \[\sum_{t=1}^T \sigma^f_{t}(x_t) \leq \sqrt{T\sum_{t=1}^T(\sigma_{t}^f(x_t))^2},\]
From Lemma 3 in \cite{ChoSayGop_17}, we have
\[\gamma^f_t \geq \frac{1}{2} \sum_{s=1}^t \ln{(1+ \lambda^{-1}(\sigma^f_{s}(x_s)))^2}\]
Combine these facts and $a \leq 2\ln(1+a),~\forall a > 0,$ we have
\begin{align*}
    \sum_{t=1}^T \sigma^f_{t}(x_t) 
    \leq &\sqrt{2T\lambda\sum_{t=1}^T \ln{(1+\lambda^{-1}(\sigma^f_{t}(x_t))^2}} \\
    \leq &\sqrt{4T\lambda \gamma^f_T}.
\end{align*}
Recall that $\beta_t^f$ is increasing with time step $t$ and $\lambda = 1 + 2/T.$ We have
\begin{align*}
    \sum_{t=1}^T\beta_t^f \sigma_t^f(x_t) \leq \beta_T^f\sqrt{4(T+2) \gamma_T^f}.
\end{align*}
Similarly, we also have
\begin{align*}
    \sum_{t=1}^T\beta_t^g \sigma_t^g(x_t) \leq \beta_T^g\sqrt{4(T+2) \gamma_T^g}.
\end{align*}
Therefore, we complete the proof.

\subsection{Proving Theorem \ref{thm: GP-UCB}} 
From Lemma \ref{lem: GP-UCB}, we have justified Condition \ref{cond} for Theorem \ref{thm: main}. Specifically, we have $\rho = 1$, $e_t^f(x) = 2\beta^f_t \sigma^f_t(x),$ and $e_t^g(x) =2\beta^g_t\sigma^g_t(x)$ in Condition \ref{cond}. From Lemma \ref{lem: GP-UCB-error}, we establish the bounds of $\sum_{t=1}^T e^f_t(x_t)$ and $\sum_{t=1}^T e^g_t(x_t)$. Now we invoke Theorem \ref{thm: main} to have 
\begin{align*}
\mathcal R(T) &= 2\beta_T^f\sqrt{4(T+2) \gamma_T^f} , \\
\mathcal V(T) &= 2\beta_T^f\sqrt{4(T+2) \gamma_T^f} + 2\beta_T^g\sqrt{4(T+2) \gamma_T^g} + 4 B_f \sqrt{T},
\end{align*} 
which completes the proof. 

\section{Proof of Theorem \ref{thm: delayGP}} \label{app: thm3}
In this section, we study SCBwC with delayed feedback, and we establish regret and violation bounds for RPOL-CensoredUCB. The detailed algorithm of RPOL-CensoredUCB is shown as follows.
\vspace{0.1in}
\hrule
\vspace{0.1in}
\noindent{\bf RPOL-CensoredUCB for SCBwC with Delayed Feedback}
\vspace{0.1in}
\hrule
\vspace{0.1in}

\noindent {\bf Initialization:}  $\mu^f_1(x) = \mu^g_1(x) = 0$, $\sigma^f_1(x) = \sigma^g_1(x) = 1$, $\forall x$, $Q_1 = 1$, $\eta_t=\sqrt{t},$ $v_t^f$ and $v_t^g$.

\noindent For $t=1,\cdots, T,$ 
\begin{itemize}
\item {\bf Pessimistic-optimistic learning:} estimate the reward $\hat f_t(x)$ and the cost $\check{g}_t(x)$ 
with GP-UCB/LCB. 
\begin{align*}
    \hat f_t(x) = \mu^f_{t}(x) + v^f_t \sigma^f_{t}(x),~\check g_t(x) = \mu^g_{t}(x) - v^g_t \sigma^g_{t}(x)
\end{align*}
\item {\bf Rectified penalty-based decision:} choose $x_{t}$ such that 
\begin{align*}
x_{t} = \argmax_{x \in \mathcal X} ~ \hat f_{t}(x) - Q_t  \check{g}^{+}_t (x)
\end{align*}
\item {\bf Feedback:} noisy delayed rewards $r_{s}(x_s)$ and constraints $c_{l}(x_l)$ revealed at time $t,$ i.e., $t=s+d^f_s$ and $t=l+d^g_l$.
\item {\bf Rectified cumulative penalty update:} 
\begin{align*}
  Q_{t+1} = \max(Q_t + \sum_{l\in \mathcal L}{c}_l^+(x_l), \eta_t), 
\end{align*}
where $\mathcal L = \{l\in [T] ~|~ t=l+d^g_l\}.$
\item{\bf Posterior model update:} update $(\mu^f_{t+1},\sigma^f_{t+1})$ and $(\mu^g_{t+1},\sigma^g_{t+1})$ with censored feedback $\Tilde{r}_{1:t+1}$ and $\Tilde{c}_{1:t+1}$:
\begin{align*}
    &\mu^f_{t+1}(x) = k^f_{t+1}(x)^T(V_{t+1}^f(\lambda))^{-1}\Tilde{r}_{1:t+1},~\mu^g_t(x) = k^g_{t+1}(x)^T(V_{t+1}^g(\lambda))^{-1}\Tilde{c}_{1:t+1}, \\
    &k^f_{t+1}(x,x^{\prime}) = k^f(x,x^{\prime}) - k^f_{t+1}(x)^T(V_{t+1}^f(\lambda))^{-1}k^f_{t+1}(x^{\prime}),\\&k^g_{t+1}(x,x^{\prime}) = k^g(x,x^{\prime}) - k^g_{t+1}(x)^T(V_{t+1}^g(\lambda))^{-1}k^g_{t+1}(x^{\prime}), \\
    &\sigma_{t+1}^f(x) = \sqrt{k_{t+1}^f(x,x)},~ \sigma_{t+1}^g(x) = \sqrt{k_{t+1}^g(x,x)},
\end{align*}
where $K^f_{t+1} := [k^f(x,x^{\prime})]_{x,x^{\prime} \in \{x_1,\cdots,x_{t}\}},$ and $V_{t+1}^f(\lambda):= K^f_{t+1} + \lambda I,$ $\Tilde{r}_{1:t+1}=[\Tilde{r}_{1,t},\cdots,\Tilde{r}_{t,t}]$, $k^f_{t+1}(x) := [k^f(x_1,x),\cdots,k^f(x_{t},x)]^T,$ $K^g_{t+1} := [k^g(x,x^{\prime})]_{x,x^{\prime} \in \{x,\cdots,x_{t}\}},$ and $V_{t+1}^g(\lambda):= K^g_{t+1} + \lambda I,$ $\Tilde{c}_{1:t+1}=[\Tilde{c}_{1,t},\cdots,\Tilde{c}_{t,t}]$, $k^g_{t+1}(x) := [k^g(x_1,x),\cdots,k^g(x_{t},x)]^T,$ $\Tilde{r}_{s,t+1} := r_s \mathbb{I}\{d^f_s \leq \min (m,t+1-s)\}$ and $\Tilde{c}_{s,{t+1}} := c_s \mathbb{I}\{d^g_s \leq \min (m,t+1-s)\}$.
\end{itemize}
\vspace{0.1in}
\hrule
\vspace{0.1in}
\subsection{Proof of Lemma \ref{lem: GP-UCB-delay}} \label{app: lem4}
Similar to Lemma \ref{lem: GP-UCB}, we justify the results for reward function $f$ and that for the constraint function $g$ follows. We first perform our analysis based on the event $\mathcal E := \{r_t \leq B_r, c_t \leq B_c, \forall t \in [T]\}$. 
Conditional on $\mathcal E,$ we have 
\begin{align*}
    \Phi_t^T \Tilde{r}_{1:t} &= \sum_{s=1}^{t-1} \varphi(x_s) r_s \mathbb{I}\{d^f_s \leq \min (m,t-s)\} \\&= \sum_{s=1}^{t-1}\varphi(x_s)r_s\mathbb{I}\{d^f_s \leq m\} + \sum_{s=t-m}^{t-1}\varphi(x_s)r_s(\mathbb{I}\{d_s^f \leq t-s\} - \mathbb{I}\{d_s^f \leq m\})
\end{align*}
Recall $V_t^f(\lambda) = (K_t^f + \lambda I)$, then we have
\begin{align*}
    \hat f(x) - \rho_m f(x) =& \mu_t^f(x) - \rho_m f(x) + v_t^f \sigma^f_t(x) \\
    =& k^f_t(x)^T(K_t^f + \lambda I)^{-1}\Tilde{r}_{1:t} - \rho_m \varphi^T(x)f + v_t^f \sigma^f_t(x) \\
    =& \varphi(x)^T \Phi_t V_t^f(\lambda)^{-1}\Tilde{r}_{1:t}- \rho_m \varphi^T(x)f + v_t^f \sigma^f_t(x) \\
    =& \varphi(x)^T  V_t^f(\lambda)^{-1}\Phi_t^T\Tilde{r}_{1:t}- \rho_m \varphi^T(x)f + v_t^f \sigma^f_t(x) \\
    =& \varphi(x)^T  V_t^f(\lambda)^{-1}(\sum_{s=1}^{t-1}\varphi(x_s)r_s\mathbb{I}\{d^f_s \leq m\}) - \rho_m \varphi^T(x)f \\
    & + \varphi(x)^T  V_t^f(\lambda)^{-1}\sum_{s=t-m}^{t-1}\varphi(x_s)r_s(\mathbb{I}\{d_s^f \leq t-s\} - \mathbb{I}\{d_s^f \leq m\}) \\
    & + v_t^f \sigma^f_t(x).
\end{align*}
From Eq.(4) in \cite{VerDaiLow_22}, we show that the following inequality hold with probability at least $1 - p/4$,
\begin{align*}
    |\varphi(x)^T  V_t^f(\lambda)^{-1}\sum_{s=t-m}^{t-1}\varphi(x_s)r_s(\mathbb{I}\{d_s^f \leq t-s\} - \mathbb{I}\{d_s^f \leq m\}) | \leq B_r \sigma_t(x)\sum_{s=t-m}^{t-1}\sigma_t(x_s).
\end{align*}
From Eq.(6) and Eq.(9) in \cite{VerDaiLow_22}, with the probability at least $1 - p/4$, we have for all $t$ such that
\begin{align*}
    |V_t^f(\lambda)^{-1}(\sum_{s=1}^{t-1}\varphi(x_s)r_s\mathbb{I}\{d^f_s \leq m\}) - \rho_m \varphi^T(x)f | \leq (B_f + (R_f + B_r)\sqrt{2(\gamma_t^f + 1 + \ln(4/p))})\sigma^f_t(x).
\end{align*}
 Recall $v^f_t = B_r\sum_{s=t-m}^{t-1}\sigma^f_{t}(x_s)+\beta^f_t$ and $\beta^f_t = B_f + (R_f+B_r)\sqrt{2(\gamma^f_{t}+1+\ln{(4/p)})}$, we prove that
 $$0 \leq \hat f_t(x) -  \rho^f_m f(x) \leq 2v^f_t \sigma^f_{t}(x), \forall t\in[T],$$ holds with the probability at least $1-p/2$. Similarly, $$0 \leq \rho^g_m g_t(x) - \check g_t(x)  \leq 2v^g_t \sigma^g_{t}(x), \forall t\in[T],$$ holds with the probability at least $1-p/2$.
 
Next, we consider the complement of the event $\mathcal E,$ i.e., $\mathcal{\bar E}.$ According to sub-Gaussian properties in Assumption \ref{assumption: noise}, we have for all $t\in [T]$
\begin{align*}
    \mathbb{P}(d^f_t\geq d_f) \leq e^{-\frac{d_f^2 }{2R_f^2}},~\mathbb{P}(d^g_t\geq d_g) \leq e^{-\frac{d_g^2 }{2R_g^2}}.
\end{align*}
Therefore, we choose $B_r = B_f + R_f \sqrt{2\log T}$ and $B_c =  B_g + R_g \sqrt{2\log T}$ such that $\mathbb P( \mathcal{\bar E}) \leq 2/T$. 

By combining the analysis on the two events $\mathcal E$ and $\mathcal {\bar E},$ we conclude that Lemma \ref{lem: GP-UCB-delay} holds with probability at least $1 - p - 2/T$ according to the union bound. 

\subsection{Proof of Lemma \ref{lem: GP-UCB-delay-error}} \label{app: lem5}
Similar to Lemma \ref{lem: GP-UCB-error},
we prove the first inequality on reward function $f$ in Lemma \ref{lem: GP-UCB-delay-error} and the second inequality on $g$ holds by following the same steps.
Recall that $v^f_t = B_r\sum_{s=t-m}^{t-1}\sigma^f_{t-1}(x_s)+\beta^f_t$, we have
\begin{align}
    \sum_{t=1}^T v_t^f \sigma_t^f(x_t) = \sum_{t=1}^T \beta_t^f \sigma_{t}^f(x_t) + \sum_{t=1}^T \sigma_{t}^f(x_t) (B_r\sum_{s=t-m}^{t-1}\sigma_{t}(x_s)). \label{items}
\end{align}
For the first term in \eqref{items}, we establish the following inequality by Lemma \ref{lem: GP-UCB-error}
\begin{align*}
    \sum_{t=1}^T \beta_t^f \sigma_t^f(x) \leq \beta_T^f \sqrt{4T\lambda \gamma^f_T}.
\end{align*}
For the second term in \eqref{items}, we have the following analysis
\begin{align*}
    \sum_{t=1}^T \sigma_{t}^f(x) (B_r\sum_{s=t-m}^{t-1}\sigma_{t}(x_s)) &= B_r \sum_{t=1}^T \sum_{s= t-m}^{t-1} \sigma_{t}^f(x_t)\sigma_{t}^f(x_s) \\
    &\leq \frac{B_r}{2} \sum_{t=1}^T \sum_{s= t-m}^{t-1} (\sigma_t^f(x_t)^2 + \sigma_t^f(x_s)^2) \\
    &\leq \frac{B_r}{2}\sum_{t=1}^T \sum_{s= t-m}^{t-1} (\sigma_t^f(x_t)^2 + \sigma_s^f(x_s)^2) \\
    &\leq m B_r \sum_{t=1}^T \sigma_t^f(x_t)^2 \\
    &\leq m B_r 4\lambda \gamma_T^f,
\end{align*}
where the last inequality holds since $\sum_{t=1}^T \sigma_t^f(x_t)^2 \leq \lambda\sum_{t=1}^T 2\ln{(1+\lambda^{-1}(\sigma^f_{t}(x_t))^2)} \leq 4\lambda \gamma_T^f$.\\
Then we have $$\sum_{t=1}^T v_t^f \sigma_t^f(x) \leq \beta_T^f \sqrt{4T\lambda \gamma^f_T} +  m B_r 4\lambda \gamma_T^f.$$
Similarly, we have the inequality for $g$ function
$$\sum_{t=1}^T v_t^g \sigma_t^g(x) \leq \beta_T^g \sqrt{4T\lambda \gamma^g_T} +  m B_c 4\lambda \gamma_T^g.$$
Therefore, we complete the proof. 

\subsection{Proving Theorem \ref{thm: delayGP}}
From Lemma \ref{lem: GP-UCB-delay}, we have justified Condition \ref{cond} for Theorem \ref{thm: main}. Specifically, we have $\rho = \rho_m$, $e_t^f(x) =2 v^f_t \sigma^f_t(x)$ and $e_t^g(x_t) = 2v^g_t\sigma^g_t(x)$ in Condition \ref{cond}. From Lemma \ref{lem: GP-UCB-delay-error}, we establish the bounds of $\sum_{t=1}^T e^f_t(x_t)$ and $\sum_{t=1}^T e^g_t(x_t)$. Now we invoke Theorem \ref{thm: main} and prove 
\begin{align*}
&\mathcal R(T) = 2\beta_T^f \sqrt{4T\lambda \gamma^f_T} + m B_r 8\lambda \gamma_T^f, \\ 
&\mathcal V(T) = 2\beta_T^f \sqrt{4T\lambda \gamma^f_T} + m B_r 8\lambda \gamma_T^f + 2\beta_T^g \sqrt{4T\lambda \gamma^g_T} + m B_c 8\lambda \gamma_T^g + 4B_f \sqrt{T}.
\end{align*} 

\section{Proof of Theorem \ref{thm: dynamicGP}} \label{app: thm4}
In this section, we study SCBwC under non-stationary environment. We establish regret and violation bounds for RPOL-SWUCB. The detailed algorithm of RPOL-SWUCB is shown as follows.
\vspace{0.1in}
\hrule
\vspace{0.1in}
\noindent{\bf RPOL-SWUCB for SCBwC under Non-stationary Environment}
\vspace{0.1in}
\hrule
\vspace{0.1in}

\noindent {\bf Initialization:} $\mu^f_1(x) = \mu^g_1(x) = 0$, $\sigma^f_1(x) = \sigma^g_1(x) = 1$, $\forall x$, $Q_1 = 1$, window size $W$, $\eta_t=\sqrt{t},$ $\beta_t^f$, $\beta_t^g$, $\Gamma_t^f$ and $\Gamma_t^g$.

\noindent For $t=1,\cdots, T,$ 
\begin{itemize}
\item {\bf Pessimistic-optimistic learning:} estimate the reward $\hat f_t(x)$ and the cost $\check{g}_t(x)$ with sliding window GP-UCB/LCB. 
\begin{align*}
    \hat f_t(x) = \mu^f_{t}(x) + \beta^f_t \sigma^f_{t}(x) + \Gamma_t^f,~\check g_t(x) = \mu^g_{t}(x) - \beta^g_t \sigma^g_{t}(x) - \Gamma_t^g.
\end{align*}
\item {\bf Rectified penalty-based decision:} choose $x_{t}$ such that \begin{align*}
x_{t} = \argmax_{x \in \mathcal X} ~ \hat f_{t}(x) - Q_t  \check{g}^{+}_t (x)
\end{align*}
\item {\bf Feedback:} noisy reward $r_{t}(x_t)$ and cost $c_{t}(x_t).$ 
\item {\bf Rectified penalty update:} 
\begin{align}
  Q_{t+1} = \max\left(Q_t + c_t^+(x_t), \eta_t\right). \nonumber
 \end{align}
\item{\bf Posterior model update:} update $(\mu^f_{t+1},\sigma^f_{t+1})$ and $(\mu^g_{t+1},\sigma^g_{t+1})$ with $r_{t_0:t+1}$ and $c_{t_0:t+1}$, where $t_0 = 1 \vee (t-W)$:
\begin{align*}
    &\mu^f_{t+1}(x) = k^f_{t+1}(x)^T(V_{t+1}^f(\lambda))^{-1}r_{t_0:t+1},~\mu^g_{t+1}(x) = k^g_{t+1}(x)^T(V_{t+1}^g(\lambda))^{-1}c_{t_0:t+1}, \\
    &k^f_{t+1}(x,x^{\prime}) = k^f(x,x^{\prime}) - k^f_{t+1}(x)^T(V_{t+1}^f(\lambda))^{-1}k^f_{t+1}(x^{\prime}),\\&k^g_{t+1}(x,x^{\prime}) = k^g(x,x^{\prime}) - k^g_{t+1}(x)^T(V_{t+1}^g(\lambda))^{-1}k^g_{t+1}(x^{\prime}), \\
    &\sigma_{t+1}^f(x) = \sqrt{k_{t+1}^f(x,x)},~ \sigma_{t+1}^g(x) = \sqrt{k_{t+1}^g(x,x)},
\end{align*}
where $k^f_{t+1}(x) := [k^f(x_{t_0},x),\cdots,k^f(x_{t},x)]^T,$ $K^f_{t+1} := [k^f(x,x^{\prime})]_{x,x^{\prime} \in \{x_{t_0},\cdots,x_{t}\}},$ and $V_{t+1}^f(\lambda):= K^f_{t+1} + \lambda I;$ $k^g_{t+1}(x) := [k^g(x_{t_0},x),\cdots,k^g(x_{t},x)]^T,$ $K^g_{t+1} := [k^g(x,x^{\prime})]_{x,x^{\prime} \in \{x_{t_0},\cdots,x_{t}\}},$ and $V_{t+1}^g(\lambda):= K^g_{t+1} + \lambda I.$
\end{itemize}
\vspace{0.1in}
\hrule
\vspace{0.1in}
\subsection{Proof of Lemma \ref{lem: GP-UCB-SW}} \label{app: lem6}
With a bit abuse of notion, we define $\Phi_t = [\varphi(x_{t_0})^T,\ldots,\varphi(x_{t-1})^T]^T.$ Recall $f_t(x) = f_t ^T\varphi(x)$, $\mu_t^f(x) = k^f_t(x)^T(V_t^f(\lambda))^{-1}r_{t_0:t} = \varphi(x)^T V_t^f(\lambda)^{-1}\Phi_{t}^T r_{t_0:t}$. Next, we study $\hat f_t(x)  - f(x)$ as follows
\begin{align*}
    \hat f_t(x) - f(x) =& \mu_t^f(x) - f_t(x) + \beta^f_t \sigma_t^f(x) + \Gamma_t^f \\
    =& \varphi(x)^T V_t^f(\lambda)^{-1}\Phi_{t}^T r_{t_0:t} - \varphi(x)^T f_t + \beta^f_t \sigma_t^f(x) + \Gamma_t^f \\
    =& \varphi(x)^T V_t^f(\lambda)^{-1}(\sum_{s=t_0}^{t-1}\varphi(x_s)\varphi(x_s)^T(f_s - f_t)) + \varphi(x)^T V_t^f(\lambda)^{-1} \varepsilon_{t_0:t} -\lambda \varphi^T(x) V_t^f(\lambda)^{-1} f_t\\&+ \beta^f_t \sigma_t^f(x)+ \Gamma_t^f.
\end{align*}
From Eq.(9) in \cite{ZhouShr_21}, we have
\begin{align*}
    |\varphi(x)^T V_t^f(\lambda)^{-1}(\sum_{s=t_0}^{t-1}\varphi(x_s)\varphi(x_s)^T(f_s - f_t))| \leq \frac{1}{\lambda} \sqrt{W2(1+\lambda)\gamma_T^f} \sum_{s=t_0}^{t-1}\|f_s - f_{s+1}\|_{k^f}.
\end{align*}
Then from Eq.(6) and Eq.(7) in \cite{ZhouShr_21}, we have
\begin{align*}
    &|\lambda \varphi^T(x) V_t^f(\lambda)^{-1} f_t| \leq B_f \sigma_t^f(x), \\
    &|\varphi(x)^T V_t^f(\lambda)^{-1} \varepsilon_{t_0:t}| \leq \frac{1}{\sqrt{\lambda}} R_f \sqrt{2\gamma^f_{t-t_0} + 2\ln(2T/p)}.
\end{align*}
Recall the value of $\beta_t^f$ and $\Gamma_t^f$, we prove that with the probability at least $1 - p/2$,
\begin{align*}
     0 \leq \hat f_{t}(x) - f_t(x) \leq 2\Gamma^f_t + 2\beta^f_t \sigma^f_{t}(x), \forall t\in[T].
\end{align*}
Similarly, $$0 \leq g_t(x) - \check g_t(x)  \leq 2\Gamma^g_t + 2\beta^g_t \sigma^g_{t}(x), \forall t\in[T]$$ holds with the probability at least $1-p/2$. Therefore, we have completed the proof according to the union bound.

\subsection{Proof of Lemma \ref{lem: GP-UCB-SW-error}} \label{app: lem7}
We first justify the upper bounds of $\sum_{t=1}^T \Gamma^f_t$  and $\sum_{t=1}^T \beta^f_t \sigma^f_{t}(x_t)$ for the reward function $f$ and the analysis for the constraint function $g$ follows the exact steps. 
Let $C_f := \frac{1}{\lambda} \sqrt{W2(1+\lambda)\gamma^f_T}.$ We have
\begin{align*}
    \sum_{t=1}^T\frac{1}{\lambda} \sqrt{2W(1+\lambda)\gamma^f_T} \sum_{s = t_0}^{t-1}\|f_s - f_{s+1}\|_{k^f} =  C_f \sum_{t=1}^T \sum_{s = t_0}^{t-1}\|f_s - f_{s+1}\|_{k^f},
\end{align*}
where $t_0 = 1\vee(t-W)$.
Combine with the fact that $\sum_{t=1}^T \sum_{s = t_0}^{t-1}\|f_s - f_{s+1}\|_{k^f} \leq W P_T$ and recall the definition of $P_T = \max(\sum_{t=1}^T\|f_{t+1}-f_t\|_{k^f},\sum_{t=1}^T\|g_{t+1}-g_t\|_{k^g}),$ we establish $$C_f \sum_{t=1}^T \sum_{s = t_0}^{t-1}\|f_s - f_{s+1}\|_{k^f} \leq C_f W P_T.$$ 
Next, we provide the bound of $\sum_{t=1}^T\beta^f_t \sigma^f_{t}(x_t).$ Note we have $$\sum_{t=1}^T\beta^f_t \sigma^f_{t}(x_t) = \sum_{t=1}^T\beta^f_t 2\sqrt{\lambda}\|\varphi(x_t)\|_{(K_{t_0:t}^f + \lambda I)^{-1}},$$ where $\varphi(\cdot)$ is the same as we define in Appendix \ref{app: lemma_2} and $K_{t_0:t}^f = \sum_{s = t_0}^{t-1} \varphi(x_s) \varphi(x_s)^T$.
\begin{align*}
    \sum_{t=1}^T\beta^f_t 2\sqrt{\lambda}\|\varphi(x_t)\|_{(K_{t_0,t}^f + \lambda I)^{-1}} \leq \beta_T^f 2\sqrt{\lambda} \sum_{k=0}^{T/W - 1}\sum_{t=kW+1}^{(k+1)W}\|\varphi(x_t)\|_{(K_{t_0,t}^f + \lambda I)^{-1}}
\end{align*}
We define $K^f_{kW+1:t} = \sum_{s=kW+1}^{t-1} \varphi(x_s)\varphi(x_s)^T$, then for $t \in [kW,(k+1)W]$. We have $(K_{t_0:t}^f + \lambda I)^{-1} \preceq (K^f_{kW+1:t} + \lambda I)^{-1}$ such that
\begin{align*}\sum_{t=kW+1}^{(k+1)W}\|\varphi(x_t)\|_{(K_{t_0:t}^f + \lambda I)^{-1}} \leq \sum_{t=kW+1}^{(k+1)W}\|\varphi(x_t)\|_{(K^f_{kW+1:t} + \lambda I)^{-1}}.
\end{align*}
Now we study every individual block (e.g., $k$-block ranges from $kW+1$ to $(k+1)W$) separately and use Lemma \ref{lem: GP-UCB-error} to conclude 
\begin{align*}
\sum_{t=kW+1}^{(k+1)W}\|\varphi(x_t)\|_{(K^f_{kW+1:t} + \lambda I)^{-1}} = \frac{1}{\sqrt{\lambda}}\sum_{t=kW+1}^{(k+1)W} \sigma^f_{t-kW}(x_t) \leq \sqrt{4T\gamma^f_T},
\end{align*}
Combine all these facts, then we prove that
\begin{align*}
    \sum_{t=1}^T \beta^f_t \sigma^f_{t}(x_t) \leq  (T/W) \beta_T^f \sqrt{4 W  \gamma^f_T} = \beta_T^f T \sqrt{\frac{4 \gamma_T^f}{W}}.
\end{align*}

\subsection{Proving Theorem \ref{thm: dynamicGP}}
Lemmas \ref{lem: GP-UCB-SW} and \ref{lem: GP-UCB-SW-error} establish the following error bounds hold with probability at least $1-p$
\begin{align*}
\sum_{t=1}^T \Gamma^f_t + \beta^f_t \sigma^f_{t}(x_t) \leq C_f W P_T + \beta_T^f T\sqrt{\frac{4 \gamma_T^f}{W}}, \\
\sum_{t=1}^T \Gamma^g_t + \beta^f_t \sigma^g_{t}(x_t) \leq C_g W P_T + \beta_T^g T\sqrt{\frac{4 \gamma_T^g}{W}}.
\end{align*}
Recall $C_f = \frac{1}{\lambda} \sqrt{W2(1+\lambda)\gamma^f_T}$ and $C_g = \frac{1}{\lambda} \sqrt{W2(1+\lambda)\gamma^g_T}$. By invoking Theorem \ref{thm: main}, we have 
\begin{align*}
&\mathcal R(T) =2C_f W P_T + 2\beta_T^f T\sqrt{\frac{4 \gamma_T^f}{W}} + 4B_f\sqrt{T}, \\ 
&\mathcal V(T) = 2C_g W P_T + 2\beta_T^g T\sqrt{\frac{4 \gamma_T^g}{W}}+ 2C_f W P_T + 2\beta_T^f T\sqrt{\frac{4 \gamma_T^f}{W}}+ 4B_f\sqrt{T}.
\end{align*} 
Let the window size $W = \gamma_T^{1/4}(T/P_T)^{1/2}$ and we have 
\begin{align*}
\mathcal R(T) \leq& 4(1+\beta^f_T) \gamma_T^{3/4} P_T^{1/4}T^{3/4} + 4B_f\sqrt{T}, \\ 
\mathcal V(T) \leq& 4(2+\beta^f_T+\beta^g_T) \gamma_T^{3/4} P_T^{1/4}T^{3/4} + 4B_f\sqrt{T}. 
\end{align*} holds with the probability at least $1-p.$ The proof is completed.  
\end{document}